\documentclass{article}

\usepackage{microtype}
\usepackage{graphicx}
\usepackage{subcaption}
\usepackage{booktabs} % for professional tables

\usepackage{hyperref}

\usepackage[accepted]{icml2026}

\usepackage{amsmath}
\usepackage{amssymb}
\usepackage{mathtools}
\usepackage{amsthm}

\usepackage[capitalize,noabbrev]{cleveref}

\theoremstyle{plain}

\theoremstyle{definition}

\theoremstyle{remark}

\usepackage{enumitem}
\setlist{nosep}

\usepackage{xspace}
 
\usepackage{array}
\usepackage{marvosym}

\usepackage{makecell}

\usepackage{multicol}

\usepackage{fontawesome5} % Added for professional IT icons

\usepackage{tikz}
  \usetikzlibrary{arrows}
  \usetikzlibrary{calc}
  \usetikzlibrary{fit}
  \usetikzlibrary {shapes.symbols}
\usetikzlibrary{shapes.geometric, arrows.meta, positioning, fit, shadows, decorations.pathreplacing, backgrounds, calc}
\usepackage{tikzpeople}

\usepackage{array}
\usepackage{multirow}

\usepackage{soul}

\usepackage{tikz-cd}
\usepackage{pgfplots}
\pgfplotsset{compat=1.17}

\usepackage{colortbl}

\icmltitlerunning{Probing Memorization of Tabular In-Context Learning} % via Membership Inference Attacks}

\begin{document}

%%%%%%%%%%%%%%%%%%%%%%%%%%%%%%%%%%%%
%%%%%%%%        Macros      %%%%%%%%
%%%%%%%%%%%%%%%%%%%%%%%%%%%%%%%%%%%%

%%%%%%%%%%%%%%%%%%%%%%%%%%%%%%%%%%%
%%%%%%%%        Macros      %%%%%%%%
%%%%%%%%%%%%%%%%%%%%%%%%%%%%%%%%%%%%
% Paragraph style for PETS
%\renewcommand{\paragraph}[1]{\subsubsection*{#1}}
% Paragraph style for usenix
%\let\oldparagraph\paragraph %save \do command in \olddo
%\renewcommand{\paragraph}[1]{\vspace{0.5\baselineskip}\textbf{#1.}}
\newcommand{\para}[1]{\vspace{0.001\baselineskip}\textbf{#1.}}

\newcommand{\iclmia}{\textsc{IclMem}\xspace}
\newcommand{\tabfms}{LTMs\xspace}
\newcommand{\tabfm}{LTM\xspace}

% Definition not defined in usenix
%\theoremstyle{definition}
%\newtheorem{definition}{Definition}[section]
%\newtheorem{lemma}{Lemma}[section]

% Tabular Cell with newline
\newcommand{\specialcell}[2][c]{%
  \begin{tabular}[#1]{@{}c@{}}#2\end{tabular}}

% cleverec for reference to sec, tab, etc
\newcommand{\todo}[1]{{\color{red}[TODO: #1]}}
\newcommand{\tocode}[1]{{\color{blue}[TOCODE: #1]}}

\newcommand{\question}[1]{{\color{orange!90!black}[Q: #1]}}
\newcommand{\answer}[1]{{\color{green!70!black}[A: #1]}}

\newcommand{\xx}[1]{{\iffalse\color{red}#1\fi}}
\newcommand{\xxx}[1]{{\color{red}#1}}
\newcommand{\tocheck}[1]{{\color{gray}#1}}
\newcommand{\tocheckhack}[1]{{\color{black}#1}}

\newcommand{\enx}[1]{\ensuremath{#1}\xspace}
\newcommand{\eps}{\ensuremath{\epsilon}\xspace}

\newcommand{\aggregator}{aggregator\xspace}

\newcommand{\ttp}{Trusted Third Party\xspace}
\newcommand{\ttps}{Trusted Third Parties\xspace}

\newcommand{\adv}{\textit{A}\xspace}
\newcommand{\adversary}{adversary\xspace}

% Define \ss command for secret sharing
\newcommand{\sscheme}{\enx{\mathsf{SS}}}

\newcommand{\tnss}{\ensuremath{(r,n)}-SS\xspace}
\newcommand{\tnssl}{\ensuremath{(r,n)}-secret sharing\xspace}

\newcommand{\nnss}{\ensuremath{(n,n)}-SS\xspace}
\newcommand{\nnssl}{\ensuremath{(n,n)}-secret sharing\xspace}

\newcommand{\sss}{Shamir's secret sharing\xspace}
\newcommand{\tss}{threshold secret sharing\xspace}

\newcommand{\ot}{OT\xspace}
\newcommand{\otl}{outsourced training\xspace}

\newcommand{\fl}{FL\xspace}

\newcommand{\gc}{{GC}\xspace}
\newcommand{\gcl}{garbled circuits\xspace}
\newcommand{\gcls}{garbled circuits\xspace}

\newcommand{\yao}{Yao's GC\xspace}
\newcommand{\yaol}{Yao's Garbled Circuits\xspace}

% numbers/count of parties
\newcommand{\nclients}{\enx{n}}
\newcommand{\nservers}{\enx{m}}

% Computing Parties
\newcommand{\cp}[1]{\enx{{CP_{#1}}}}
\newcommand{\cpl}{computing party\xspace}
\newcommand{\cpls}{computing parties\xspace}

% Server
\newcommand{\s}[1]{\enx{{S_{#1}}}}
\newcommand{\slong}{server\xspace}
\newcommand{\sln}[1]{server {#1}\xspace}
\newcommand{\sls}{servers\xspace}

% Parties
\newcommand{\p}[1]{\enx{{P_{#1}}}}
\newcommand{\pl}{party\xspace}
\newcommand{\pln}[1]{\enx{party_{#1}}}
\newcommand{\pls}{parties\xspace}

% Input Parties
\newcommand{\ip}[1]{\enx{{IP_{#1}}}}
\newcommand{\ipl}{input party\xspace}
\newcommand{\ipln}[1]{\mathrm{input party {#1}}}
\newcommand{\ipls}{input parties\xspace}

\let\oldc\c
\renewcommand{\c}[1]{\enx{{C_{#1}}}}
\newcommand{\cl}{client\xspace}
\newcommand{\cln}[1]{client {#1}}
\newcommand{\cls}{clients\xspace}

% Noise
\newcommand{\npartial}[1]{\enx{\rho_{#1}}}%{\enx{\rho_\mathsf{part}}}
\newcommand{\ncentral}{\enx{\zeta}}

% Data Owners
\let\olddo\do %save \do command in \olddo
\renewcommand{\do}{\textit{MO}\xspace}
\newcommand{\dataowners}{data owners\xspace}
\newcommand{\dataowner}{model owner\xspace}

\newcommand{\Xmark}{$\textcolor{red}{\mathbf{\times}}$}
\newcommand{\XmarkGood}{$\textcolor{cyan!60!black}{\mathbf{\times}}$}
\newcommand{\questionmark}{$\mathbf{\textcolor{orange}{?}}$}
\newcommand{\doubledash}{$\mathbf{\textcolor{orange}{\--}}$}
\newcommand{\tickmarker}{$\mathbf{\textcolor{cyan!60!black}{\checkmark}}$}
\newcommand{\tickmarkerRed}{$\mathbf{\textcolor{red}{\checkmark}}$}

%Used to suppress the error of [H] option with algorithm which works but needs to be fixed
%\makeatletter
%\newcommand{\removelatexerror}{\let\@latex@error\@gobble}
%\makeatother

\newcommand{\perturbinput}{\enx{\mathsf{\textcolor{blue}{PerturbInput}}}}
\newcommand{\perturbloss}{\enx{\mathsf{\textcolor{red}{PerturbLoss}}}}
\newcommand{\perturbgrad}{\enx{\mathsf{\textcolor{green!60!black}{PerturbGradient}}}}
% before used textt for macros
\newcommand{\perturbout}{\enx{\mathsf{\textcolor{orange}{PerturbOutput}}}}
\newcommand{\perturblabel}{\enx{\mathsf{\textcolor{violet}{PerturbLabel}}}}

\newcommand{\pgrad}{\enx{\mathsf{\textcolor{green!60!black}{Gradient}}}}
\newcommand{\pout}{\enx{\mathsf{\textcolor{orange}{Output}}}}
\newcommand{\plabel}{\enx{\mathsf{\textcolor{violet}{Label}}}}

% Data Owners
%\let\oldparagraph\paragraph %save \do command in \olddo
%\renewcommand{\paragraph}[1]{\vspace{0.5\baselineskip}\textbf{#1.}}
%\renewcommand{\paragraph}[1]{\textbf{#1}}

\newcounter{numberedparagraph}
\newlength{\parindentoriginal}
\newenvironment{numberedparagraphs}
{
    \setcounter{numberedparagraph}{0}
    \renewcommand{\paragraph}[1]{\refstepcounter{numberedparagraph}\textbf{\thenumberedparagraph) \hspace{0.1em}##1.}}
}{
    \setcounter{numberedparagraph}{0}
}

% Macros for protocol phases
\newcommand{\phaseformat}[1]{\enx{\mathsf{#1}}}
\newcommand{\setup}{\phaseformat{Setup}}
\newcommand{\compgrad}{\phaseformat{GradientCompute}}
\newcommand{\protection}{\phaseformat{Protect}}
\newcommand{\perturb}{\phaseformat{Perturb}}
\newcommand{\aggr}{\phaseformat{Aggregate}}
\newcommand{\updatemodel}{\phaseformat{Update}}
%
%
%% Noise generation protocol
\newcommand{\centralnoise}{centralized noise sampling\xspace}
\newcommand{\partialnoise}{partial noise aggregation\xspace}
\newcommand{\mpcnoise}{distributed noise sampling\xspace}
%
%%Noise macros for algorithm
\newcommand{\partnoise}{\enx{\mathsf{\textcolor{blue!60!black}{{PNoise}}}}}
\newcommand{\centrnoise}{\enx{\mathsf{\textcolor{blue!60!black}{CNoise}}}}
\newcommand{\funcnoise}{\enx{\mathsf{\textcolor{blue!60!black}{FNoise}}}}
%
%% examples macros
\newcommand{\eg}{e.g.,\;}
\newcommand{\ie}{i.e.,\;}

% Secret sharing macro

\newcommand{\shares}[1]{\langle {#1} \rangle }
\newcommand{\ashares}[1]{\langle {#1} \rangle^A}
\newcommand{\bshares}[1]{\langle {#1} \rangle^B}

% Clipping parameter macro
\newcommand{\clipparam}{C}
\newcommand{\clip}{\enx{\mathsf{Clip}}}

% local global computations highlight
%\DeclareRobustCommand{\local}[1]{{\sethlcolor{orange}\hl{#1}}}
% local only used to align row in the comparison algorithm
\newcommand{\local}[1]{\colorbox{white}{#1}}
\newcommand{\glob}[1]{\colorbox{\mylighterblue}{#1}}

\newcommand{\privguarantee}{DP Considerations}

%%% table
\newcommand{\tabcolor}{pyblue}
\newcommand{\tabheadform}[1]{{\textsf{\color{\tabcolor}#1}}}
\newcommand{\tableheader}[1]{\tabheadform{\itshape\large#1}} %scshape
\newcommand{\tablesubheader}[1]{\tabheadform{#1}} %scshape
\newcommand{\tableemph}[1]{\tabheadform{#1}}
\newcolumntype{C}[1]{>{\centering\arraybackslash}m{#1}}

\newcommand{\theader}[1]{\textbf{\textsf{#1}}} %scshape

\newcommand{\Enc}{\enx{\mathsf{Enc}}}
\newcommand{\Dec}{\enx{\mathsf{Dec}}}
\newcommand{\Shr}{\enx{\mathsf{Shr}}}
\newcommand{\Rec}{\enx{\mathsf{Rec}}}
\newcommand{\Mask}{\enx{\mathsf{Mask}}}
\newcommand{\UnMask}{\enx{\mathsf{UnMask}}}
\newcommand{\Garble}{\enx{\mathsf{Garble}}}
\newcommand{\Eval}{\enx{\mathsf{Eval}}}

\newcommand{\LaplaceITS}{\textsf{LaplaceITS}}
\newcommand{\DiscLaplace}{\textsf{DiscLaplace}}
\newcommand{\DiscGaussian}{\textsf{DiscGauss}}
\newcommand{\Skellam}{\textsf{Skellam}}
\newcommand{\BoxMuller}{\textsf{BoxMuller}}
\newcommand{\Poisson}{\textsf{Poisson}}
\newcommand{\Bern}{\textsf{Bern}}
\newcommand{\Geom}{\textsf{Geom}}

%% Observation counter
\newcounter{obscounter}
\setcounter{obscounter}{1}

\newcommand{\obs}[1]{%
  \textbf{(O\theobscounter)~{#1}}%
  \stepcounter{obscounter}%
}

\newcommand{\obsmanual}[2]{\textbf{(O#1)~{#2}}}
\newcommand{\obsn}[1]{\textbf{(O#1)}}

\newcounter{researchdirectioncounter}
\setcounter{researchdirectioncounter}{1}

\newcommand{\research}[1]{%
  \textbf{(D\theresearchdirectioncounter)~{#1}.}%
  \stepcounter{researchdirectioncounter}%
}

\newcommand{\researchmanual}[2]{\textbf{(D#1)~{#2}}}

\newcommand{\code}[1]{\textbf{{#1}}}

\newcommand{\smax}{\enx{\mathsf{Softmax}}}
\newcommand{\smaxmpc}{\enx{\mathsf{Softmax_{\mathsf{MPC}}}}}
\newcommand{\erf}{\enx{\mathsf{erf}}}
\newcommand{\smaxmax}{\enx{\mathsf{Softmax{\capconst}}}}
\newcommand{\gelu}{\enx{\mathsf{GeLU}}}
\newcommand{\gelubolt}{\enx{\mathsf{GeLU_{\mathsf{MPC}}}}}
\newcommand{\gelumpc}{\enx{\mathsf{GeLU_{\mathsf{MPC}}}}}
\newcommand{\geluoptim}{\enx{\mathsf{GeLU_{MPC}}}}
\newcommand{\relu}{\enx{\mathsf{ReLU}}}
\newcommand{\softcap}{\enx{\mathsf{SoftCap}}}
\newcommand{\layernorm}{\enx{\mathsf{LayerNorm}}}
\newcommand{\layernormmpc}{\enx{\mathsf{LayerNorm_{\mathsf{MPC}}}}}
\newcommand{\classifier}{\enx{\mathsf{Classifier}}}
\newcommand{\embedding}{\enx{\mathsf{Embedding}}}
\newcommand{\attention}{\enx{\mathsf{Attention}}}
\newcommand{\linear}{\enx{\mathsf{Linear}}}
\newcommand{\model}[1]{\enx{\mathcal{M}^A_{\text{#1}}}}
\newcommand{\modelplain}[1]{\enx{\mathcal{M}_{\text{#1}}}}
\newcommand{\param}[1]{\enx{\theta_{\mathrm{#1}}}}
\newcommand{\weight}[1]{\enx{\mathbf{W}_{\text{#1}}}}
\newcommand{\bias}[1]{\enx{\mathbf{b}_{\text{#1}}}}

\newcommand{\tanhf}{\enx{\mathsf{Tanh}}}
\newcommand{\invsqrt}{\enx{\mathsf{InvSqrt}}}

\newcommand{\invsqrtcrypten}{\enx{\invsqrt_{\mathsf{\crypten}}}}
\newcommand{\invsqrtmpc}{\enx{\invsqrt_{\mathsf{MPC}}}}

\newcommand{\user}{\enx{\mathcal{U}}}
\newcommand{\usr}[1]{\enx{{U_{#1}}}}

\newcommand{\capconst}{\enx{\mathsf{K}}}
\newcommand{\dataset}[1]{\enx{{D}_{\text{#1}}}}
\newcommand{\batch}[1]{\enx{\mathcal{B}_{\text{#1}}}}

\newcommand{\preparemodel}{\enx{\mathsf{AdaptModel}}}
\newcommand{\mapmpc}{\enx{\mathsf{MPCfyModel}}}
\newcommand{\finetune}{\enx{\mathsf{DPFinetune}}}
\newcommand{\inference}{\enx{\mathsf{Inference_{MPC}}}}
\newcommand{\dpadambc}{\enx{\mathsf{DPAdamBC}}}
\newcommand{\dpadamw}{\enx{\mathsf{DPAdamW}}}
\newcommand{\subsampling}{\enx{\mathsf{PoissonSampling}}}
\newcommand{\replacetorch}{\enx{\mathsf{BuildMPCModel}}}

\newcommand{\matquery}{\enx{\mathbf{Q}}}
\newcommand{\matkey}{\enx{\mathbf{K}}}
\newcommand{\matvalue}{\enx{\mathbf{V}}}
\newcommand{\matinput}{\enx{\mathbf{X}}}
\newcommand{\vecinput}{\enx{\mathbf{x}}}
\newcommand{\matoutput}{\enx{\mathbf{Y}}}
\newcommand{\vecoutput}{\enx{\mathbf{y}}}
\newcommand{\matweight}[1]{\enx{\mathbf{W}_{#1}}}
\newcommand{\loraA}{\enx{\mathbf{A}}}
\newcommand{\loraB}{\enx{\mathbf{B}}}

\newcommand{\noise}[1]{\enx{\Psi_{#1}}}
\newcommand{\matnoise}[1]{\noise{#1}}

\newcommand{\bert}{BERT}
\newcommand{\roberta}{Ro{\bert}a}
\newcommand{\gpt}{GPT}
\newcommand{\lora}{\enx{\mathsf{LoRA}}}
\newcommand{\falora}{\enx{\mathsf{FALoRA}}}
\newcommand{\loralinear}{\enx{\mathsf{FALoRALinear}}}

\newcommand{\sst}{{{SST2}}\xspace}
\newcommand{\qnli}{{{QNLI}}\xspace}
\newcommand{\mnli}{{{MNLI}}\xspace}
\newcommand{\qqp}{{{QQP}}\xspace}

\newcommand{\lr}{\enx{\eta}}
\newcommand{\batchsize}{\enx{B}}
\newcommand{\nepochs}{\enx{E}}
\newcommand{\lorarank}{\enx{r}}
\newcommand{\loraalpha}{\enx{\alpha}}

\newcommand{\ghostclip}{\enx{\mathsf{GhostClipping}}}

\newcommand{\ffn}{\enx{\mathsf{FFN}}}
\newcommand{\encoder}{\enx{\mathsf{Encoder}}}

\newcommand{\sota}{state-of-the-art}
\newcommand{\mlaas}{MLaaS\xspace}
\newcommand{\crypten}{CrypTen}

\newcommand{\precision}{\enx{{f}}}

\newcommand{\lan}{LAN\xspace}
\newcommand{\wanr}{WAN$_\text{R}$\xspace}
\newcommand{\wang}{WAN$_\text{G}$\xspace}

\newcommand{\lancond}{(3Gbps, 0.8ms)\xspace}  %(3Gbps, 0.8ms), (200Mbps, 40ms), (100Mbps,80ms)
\newcommand{\wanrcond}{(200Mbps, 40ms)\xspace}
\newcommand{\wangcond}{(100Mbps, 80ms)\xspace}

\newcommand{\kaiuni}{\enx{\mathsf{Kaiming\ Uniform}}}
\newcommand{\kainorm}{\enx{\mathsf{Kaiming\ Normal}}}
\newcommand{\xavuni}{\enx{\mathsf{Xavier\ Uniform}}}
\newcommand{\xavnorm}{\enx{\mathsf{Xavier\ Normal}}}
\newcommand{\norm}{\enx{\mathsf{Normal}}}
\newcommand{\orth}{\enx{\mathsf{Orthogonal}}}

\newcommand{\bolt}{\enx{\mathsf{BOLT}}}
\newcommand{\shaft}{\enx{\mathsf{SHAFT}}}
\newcommand{\bumblebee}{\enx{\mathsf{BumbleBee}}}
\newcommand{\iron}{\enx{\mathsf{Iron}}}
\newcommand{\sprint}{\enx{\mathsf{SPRINT}}}
\newcommand{\sprintclear}{\enx{\mathsf{SPRINT}_{\text{clear}}}}
\newcommand{\sprintmpc}{\enx{\mathsf{SPRINT}_{\text{MPC}}}}
\newcommand{\pp}{\enx{\text{\,pp}}}

\newcommand{\negvspacealgo}{\vspace{-0.57em}}

\newcommand{\securityparam}{\enx{\kappa}}

%%%%%%%%%%%%%%%%%%%%%%%%%%%%%%%%%%%%

%%%%%%%%%%%%%%%% Authors' Info %%%%%%%%%%%%%%%%%
\twocolumn[
  %\icmltitle{Towards Membership Inference Attacks of Tabular In-Context Learning Foundation Models.}
  %\icmltitle{Probing Memorization of Pre-Training Data in Tabular In-Context Learning Foundation Models via Membership Inference Attacks}
  \icmltitle{Probing Memorization of Tabular In-Context Learning} % via Membership Inference Attacks}
  %\icmltitle{Evaluating Memorization of Pre-Training Data in In-Context Learning Large Tabular Models}
  \icmlsetsymbol{equal}{*}
  % It is OKAY to include author information, even for blind submissions: the
  % style file will automatically remove it for you unless you've provided
  % the [accepted] option to the icml2026 package.

  % List of affiliations: The first argument should be a (short) identifier you
  % will use later to specify author affiliations Academic affiliations
  % should list Department, University, City, Region, Country Industry
  % affiliations should list Company, City, Region, Country

  % You can specify symbols, otherwise they are numbered in order. Ideally, you
  % should not use this facility. Affiliations will be numbered in order of
  % appearance and this is the preferred way.
  \begin{icmlauthorlist}
  \icmlauthor{Francesco Capano}{sap}
  \icmlauthor{Jonas B{\"o}hler}{sap}
  \end{icmlauthorlist}

  \icmlaffiliation{sap}{SAP SE, Walldorf, Germany}
  \icmlaffiliation{sap}{SAP SE, Walldorf, Germany}
  \icmlcorrespondingauthor{Francesco Capano}{francesco.capano@sap.com}
  \icmlcorrespondingauthor{Jonas B{\"o}hler}{jonas.boehler@sap.com}

  \icmlkeywords{Tabular FM, In-context learning, Privacy, Membership inference}

  \vskip 0.3in
]

% this must go after the closing bracket ] following \twocolumn[ ...

% This command actually creates the footnote in the first column listing the
% affiliations and the copyright notice. The command takes one argument, which
% is text to display at the start of the footnote. The \icmlEqualContribution
% command is standard text for equal contribution. Remove it (just {}) if you
% do not need this facility.

% Use ONE of the following lines. DO NOT remove the command.
% If you have no special notice, KEEP empty braces:
\printAffiliationsAndNotice{}  % no special notice (required even if empty)
% Or, if applicable, use the standard equal contribution text:
% \printAffiliationsAndNotice{\icmlEqualContribution}

\begin{abstract}
%\todo{Read suggestions for abstract and update.}
%
%Suggestion for (parts of) the abstract:
%Tabular foundation models (\tabfms) leveraging in-context learning (ICL) achieve state-of-the-art performance on tabular tasks.
%%
%\tabfms are often trained on synthetic data alleviating privacy concerns. However, more recently real-world data is used for its rich semantics {cite: RealTab, TabDP, Contexttab - please check}.
%While countless works aim to extract information about training data from LLMs, such privacy considerations for \tabfms trained on real data remain under-explored.
%%
%To close this gap, we introduce \iclmia, a framework towards auditing the privacy of tabular in-context learning with realistic data supporting regression and classification.
%At the heart of \iclmia are context-perturbing membership inference attacks (MIA) - aiming to separate desired context-based predictions from parametrical memorization.
%We carefully account for common MIA pitfalls, i.e., false positives due to distribution shift in train/test split addressed via statistical tests and uncalibrated predictions via a reference model.
%%
%Our preliminary MIA evaluations for LTM Contexttab across 26 CARTE and SALT benchmark datasets finds no significant attack signal when MIA pitfalls are factored in.
%}
Large tabular models (LTMs), i.e., tabular foundation models leveraging in-context learning (ICL), achieve state-of-the-art performance on tabular tasks. 
While LLMs are known to unintentionally memorize training data, the memorization dynamics of LTMs remain largely unexplored. 
We investigate the potential for parametric memorization in tabular ICL. % via a controlled fine-tuning setup. 
We introduce \iclmia, a probing framework designed to separate context-based predictions from parametric memorization.
Our zero-information multiple-choice context strips away valid contextual patterns to force the model to fall back on its parametric memory. 
Our controlled fine-tuning setup establishes membership ground truth and accounts for common pitfalls, e.g., distribution shift, feature contamination, base-rate fallacy, and the pre-trained base model acts as reference to calibrate for sample difficulty.
Our controlled evaluation on a leading real-world-trained \tabfm detects moderate memorization signals in 8 out of 10 tasks ($\text{AUC}$ up to  $0.67$ and TPR at $1\%$ FPR $>0.1$).
Notably, memorization signals are strongest for low-cardinality and binary tasks. However, they largely vanish under realistic training conditions.
%
%\tocheck{Under joint multi-task training with dynamic context-query sampling, detection collapses to a single dataset at $Q=512$, and even that residual signal requires more than 10 epochs — exceeding typical \tabfm pre-training budgets.}
%
Our findings show LTM memorization signals under specific circumstances (single-task fine-tuning with fixed samples across many epochs and small query size). To protect sensitive data, appropriate measures must be taken, which we discuss.

\end{abstract}

% Colors based on https://blog.datawrapper.de/beautifulcolors/
\definecolor{mylightblue} {RGB}{109,196,233}%{ 72,178,222}
\definecolor{myblue}      {RGB}{ 76,144,186}
\definecolor{mybluegreen} {RGB}{ 43,194,194}
\definecolor{mydarkgreen} {RGB}{32,149,130}%{  0,138,113}
\definecolor{mylightgreen}{RGB}{ 196, 247, 192}
\definecolor{myyellow}    {RGB}{247,212,140}
\definecolor{mycyan}      {RGB}{17, 202, 247}
\definecolor{mygreen}     {RGB}{107, 250, 169}
\definecolor{myred}       {RGB}{224, 91, 91}
\definecolor{mylightred}  {RGB}{247, 237, 236}
%\definecolor{mylightyellow}{RGB}{250,214,115}
%\definecolor{myyellow}    {RGB}{244,184, 17}
%\definecolor{myyellow}    {RGB}{254,217,142}
%\definecolor{mylightorange}{RGB}{250,159, 76} %{255,172, 88}
\definecolor{myorange}    {RGB}{254,153, 41}%{254,178, 76}
%\definecolor{myorange}    {RGB}{255,145, 43}
\definecolor{mydarkorange}{RGB}{202, 96,  0}
\definecolor{mylightred}  {RGB}{229,120,114}
\definecolor{myred}       {RGB}{240, 59, 32}
%\definecolor{myred}       {RGB}{214, 81, 36} %\definecolor{myred}      {RGB}{222,102, 62}

% colors based on Python's matplotlib
\definecolor{pyblue}      {RGB}{ 32,119,180}
\definecolor{pyorange}    {RGB}{248,118,  0}
\definecolor{pygreen}     {RGB}{ 43,153, 43}
\definecolor{pyred}       {RGB}{207, 37, 37}

% colors based on https://colorbrewer2.org/#type=sequential&scheme=YlOrBr&n=4
\definecolor{mylightblueb}{RGB}{ 65,182,196} %{67,162,202}
\definecolor{myblueb}     {RGB}{ 34, 94,168}
\definecolor{mybluegreenb}{RGB}{161,218,180}
\definecolor{myyellowb}   {RGB}{254,217,142}%{254,232,200}%{255,237,160}
\definecolor{myorangeb}   {RGB}{254,153, 41}%{254,178, 76}
\definecolor{myredb}      {RGB}{240, 59, 32}

\definecolor{mycyanbg}      {RGB}{201, 245, 252}
\definecolor{myorangebg}      {RGB}{252, 210, 131}
\definecolor{mygraybg}{gray}{0.93}

% based on above colors - requires backslash (i.e., fill=\mylighterblue, instead of fill=myblue in tikz)
\def\mylighterblue{ mylightblue!30}
  \def\mylighterbluetwo{mylightblue!15}
\def\mylighterred{  mylightred!30}
\def\mylightergreen{mylightgreen!30}
\def\mylighterorange{myorange!20}

\def\mydarkerblue{ mylightblue!85}
  \def\mydarkerbluetwo{ mylightblue!70}
\def\mydarkerred{  mylightred!85}
\def\mydarkergreen{mylightgreen!85}
\def\mydarkerorange{myorange!85}

% DP / Privacy models
\def\clientbordercolor{gray}
\def\clientfillcolor{gray!25}
\def\mpcserverfillcolor{\mylighterbluetwo}

% color evaluation \setup: AWS regions and connection (pipe) between them
\def\regionfillcolor{mylightblue!25}
\def\regionbordercolor{\mydarkerblue}
\def\pipecolor{gray!5}
\def\awsclientfillcolor{\mpcserverfillcolor}
\def\awsclientbordercolor{\clientbordercolor} %mylightblue!95!black}

% 2PC - Example Table with highlighted columns
\def\tabcolorlight{mylightblue!75}
\def\tabcolordark{gray!18}
%\definecolor{tabcolorlight}{RGB}{198,224,250}
%\definecolor{tabcolordark}{RGB}{110,165,215}
  %\definecolor{tabcolordark}{RGB}{105,180,245}

% 2PC - Accuracy Graphs: Pruned vs Original
\def\origColor{myred!85!black}
\def\prunedColor{pyblue!95!black}
%% Accuracy Graphs: Exponential Mechanism vs Smooth Sensitivity
%\def\ExpMechTwoGraphsColor{\prunedColor}
%\def\SmoothTwoGraphsColor{\origColor}

% MPC - Accuracy Graphs: Related Work
\def\ExpMechColor{pyblue!80!black}
\def\ExpMechSubrangeColor{mydarkgreen!72!white}%{mydarkgreen!95!white}
\def\SmoothSensColor{pyorange!70!white}
\def\SampleAggColor{pyred}
%%\def\ExpMechColor{mybluegreenb}
%%\def\ExpMechSubrangeColor{mylightblueb}
%%\def\SmoothSensColor{myorangeb}
%%\def\SampleAggColor{myredb}

% MPC - Privacy vs Running Time Graph
\def\leftAxisColor{\origColor}
\def\rightAxisColor{\prunedColor}

% MPC - Running Time for different domain sizes
\def\smallDomainColor{myblue}
\def\mediumDomainColor{mylightgreen}
\def\largeDomainColor{mylightred}

%ARROW
\def\arrowColor{gray!75}

%\input{todos.tex}

% Intro
\section{Introduction}

Tabular data dominate enterprise applications and domains such as healthcare and finance \cite{chui2018notes}.
%enterprise domains format in enterprise settings and domains such as healthcare and finance \cite{chui2018notes}. 
%
Large tabular models (\tabfms) \cite{van2024tabular,hollmann2022tabpfn,spinaci2025contexttab}, i.e., tabular foundation models, achieve state-of-the-art performance for tabular predictions by leveraging in-context learning (ICL) to perform zero-shot predictions on unseen tasks \cite{brown2020language}. 
At inference, the model receives a context (e.g., historical example rows) alongside a query row with a missing target field to be predicted. 
%
%This enables the model to perform zero-shot predictions on unseen tasks without the need for task-specific fine-tuning.
%
While \tabfms are typically pre-trained on synthetic data \cite{hollmann2022tabpfn,qu2025tabicl}, recent models incorporate real-world tabular corpora \cite{spinaci2025contexttab, garg2025real, ma2024tabdpt}, e.g., by fine-tuning on diverse real-world task mixtures to improve performance \cite{grinsztajn2025tabpfn}.
%
%Consequently, the risk of unintentional memorization arises.
%
While LLMs are known to unintentionally memorize training data \cite{chen2026membership,hayes2025strong, szep2026unintended}, the memorization dynamics of \tabfms remain largely unexplored.
%
%Membership inference attacks (MIAs) \cite{shokri2017membership} aide in memorization probing and aim to infer whether a specific record was part of the training data.
%
%Whether \tabfms exhibit the same memorization behavior, and whether it is detectable via MIAs, remains an open question.
Rigorous memorization probing is crucial to inform potential mitigations for sensitive fine-tuning data, such as differential privacy and adaptation of data pre-processing and training regimes.

\para{Problem Setting}
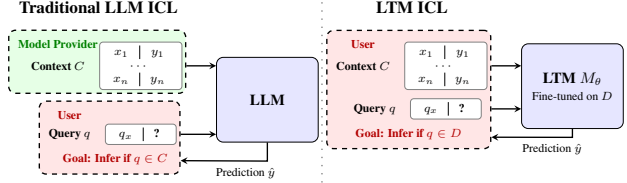
\begin{figure}[t!]
\centering
\resizebox{\columnwidth}{!}{
\begin{tikzpicture}[
    model/.style={rectangle, draw, rounded corners, minimum width=6em, minimum height=5em, align=center, fill=blue!10, font=\small},
    adv/.style={rectangle, draw, rounded corners, align=center, fill=red!10, dashed, thick, inner sep=0.2em},
    owner/.style={rectangle, draw, rounded corners, align=center, fill=green!10, dashed, thick, inner sep=0.2em},
    tablebox/.style={rectangle, draw=gray, rounded corners=0.2em, fill=white, align=center, inner sep=0.2em, font=\scriptsize},
    arrow/.style={thick, ->, >=stealth}
]

% ==========================================
% LEFT SIDE: TRADITIONAL LLM SETTING
% ==========================================
\node (llm_title) {\textbf{Traditional LLM ICL}};

% Owner Box (Provides Context)
\node[tablebox, below=1.3em of llm_title, xshift=2.5em] (llm_context_box) {
    \begin{tabular}{c|c} $x_1$ & $y_1$ \\ \multicolumn{2}{c}{$\cdots$} \\ $x_n$ & $y_n$ \end{tabular}
};
\node[left=0.5em of llm_context_box, font=\scriptsize\bfseries, align=right] (llm_context_lbl) {Context $C$};
\node[above=0.1em of llm_context_lbl, font=\scriptsize\bfseries, text=green!50!black] (owner_lbl) {Model Provider};

\begin{scope}[on background layer]
    \node[owner, fit=(owner_lbl) (llm_context_lbl) (llm_context_box)] (owner_box) {};
\end{scope}

% User Box (Provides Query)
\node[tablebox, below=2.0em of llm_context_box] (llm_query_box) {
    \begin{tabular}{c|c} $q_x$ & \textbf{?} \end{tabular}
};
\node[left=0.5em of llm_query_box, font=\scriptsize\bfseries, align=right] (llm_query_lbl) {Query $q$};
\node[above=-0.1em of llm_query_lbl, font=\scriptsize\bfseries, text=red!70!black] (adv_llm) {User};
\node[below=0.2em of llm_query_box, font=\scriptsize\bfseries, text=red!70!black, xshift=-1.5em] (llm_adv_goal) {Goal: Infer if $q \in C$};

\begin{scope}[on background layer]
    \node[adv, fit=(llm_query_box) (adv_llm) (llm_adv_goal) (llm_query_lbl)] (llm_adv_box) {};
\end{scope}

% LLM Model
\node[model, right=2em of llm_context_box, yshift=-2.0em] (llm_model) {\textbf{LLM}};

% Flows
\draw[arrow] (owner_box.east |- llm_context_box) -- (llm_model.west |- llm_context_box);
\draw[arrow] (llm_adv_box.east |- llm_query_box) -- (llm_model.west |- llm_query_box);
% Feedback Loop
\draw[arrow] (llm_model.south) -- ++(0,-1.1em) -- node[pos=0.25, below, font=\scriptsize] {Prediction $\hat{y}$} (llm_adv_box.342);

% Divider
\draw[thick, loosely dotted, gray] (llm_title.north east) ++(8.15em,0.1em) -- ++(0,-11.5em);

% ==========================================
% RIGHT SIDE: TABFM MIA SETTING
% ==========================================
\node[right=11em of llm_title] (tabfm_title) {\textbf{LTM ICL}};

% User Box (Controls BOTH Context and Query)
\node[tablebox, below=1.3em of tabfm_title, xshift=2.0em] (tab_context_box) {
    \begin{tabular}{c|c} $x_1$ & $y_1$ \\ \multicolumn{2}{c}{$\cdots$} \\ $x_n$ & $y_n$ \end{tabular}
};
\node[left=0.5em of tab_context_box, font=\scriptsize\bfseries, align=right] (tab_context_lbl) {Context $C$};

\node[tablebox, below=0.5em of tab_context_box] (tab_query_box) {
    \begin{tabular}{c|c} $q_x$ & \textbf{?} \end{tabular}
};
\node[left=0.5em of tab_query_box, font=\scriptsize\bfseries, align=right] (tab_query_lbl) {Query $q$};

\node[above=0.1em of tab_context_lbl, font=\scriptsize\bfseries, text=red!70!black] (tab_adv_lbl) {User};
\node[below=0.2em of tab_query_box, font=\scriptsize\bfseries, text=red!70!black, xshift=-2.2em, align=center] (tab_adv_goal) {Goal: Infer if $q \in D$};
%Probe memorization\\of $q$ from $D$};

\begin{scope}[on background layer]
    \node[adv, fit=(tab_adv_lbl) (tab_context_lbl) (tab_context_box) (tab_query_lbl) (tab_query_box) (tab_adv_goal)] (tab_adv_box) {};
\end{scope}

% LTM Model
\node[model, right=2em of tab_context_box, yshift=-1.2em] (tab_model) {\textbf{LTM} $M_\theta$\\ \scriptsize Fine-tuned on $D$};

% Flows
\draw[arrow] (tab_adv_box.east |- tab_context_box) -- (tab_model.west |- tab_context_box);
\draw[arrow] (tab_adv_box.east |- tab_query_box) -- (tab_model.west |- tab_query_box);
% Feedback Loop
\draw[arrow] (tab_model.south) -- ++(0,-0.5em) -- node[pos=0.25, below, font=\scriptsize] {Prediction $\hat{y}$} (tab_adv_box.330);

\end{tikzpicture}
}
\vspace{-2.0em}
\caption{Memorization probing for LLM vs LTM. \textbf{Left:} In LLMs with ICL the model provider supplies a sensitive context $C$, which the user attempts to extract via user-controlled queries. \textbf{Right:} In \tabfms, the user controls both the context $C$ and the query $q$. In \iclmia the context is used to probe whether $q$ was memorized from the sensitive fine-tuning data $D$.}
\label{fig:threat_model}
\vspace{-1.5em}
\end{figure}
In the ICL paradigm, a model predicts a missing target value $y_q$ given query's input features $x_q$, and a context $C = \{(x_1, y_1), \dots, (x_n, y_n)\}$ of examples.
With \tabfms, the user provides both the context $C$ and the query $q$, as shown in Fig.~\ref{fig:threat_model}.
We assume an \tabfm $M_\Theta$ is fine-tuned on sensitive $D$ by a data owner who provides access to the fine-tuned \tabfm to users.
This setting opens up the possibility to manipulate $C$ arbitrarily, to probe differences in model behavior on context-query pairs derived from data included in or excluded from $D$. 
In contrast, existing works on ICL privacy primarily assume the model provider defines a sensitive $C$, which an attacker attempts to extract via membership inference attacks (MIA) by manipulating $q$ \cite{wen2024membership, duan2024privacy}. 
MIAs \cite{shokri2017membership} aim to infer whether a specific record was part of the training data, and transferring MIAs to \tabfms presents unique challenges. 
Unlike LLMs, which can generate arbitrary text on a vast output space to extract memorized tokens \cite{carlini2021extracting}, \tabfms are encoder-only architectures that strictly constrain predictions to the pre-defined candidate labels provided in $C$ \cite{spinaci2025contexttab,hollmann2022tabpfn}.
%
%, \tabfms are encoder-only architectures that strictly constrain predictions to the pre-defined candidate labels provided in $C$ \cite{spinaci2025contexttab,hollmann2022tabpfn}.
%Typical privacy investigations target LLMs capable of arbitrary text generation, relying on a vast output space to extract memorized tokens \cite{carlini2021extracting}.  
%%
%However, \tabfms are encoder-only architectures that strictly constrain predictions to the pre-defined candidate labels provided in $C$ \cite{spinaci2025contexttab,hollmann2022tabpfn}. 
%
More importantly, while masked-language modeling can induce memorization \cite{hartmann2023sok}, \tabfms operate as Prior-Data Fitted Networks (PFNs) \cite{muller2021transformers}. They are trained to infer structural connections within the context rather than memorize explicit targets \cite{muller2021transformers}.
This raises a fundamental, yet unanswered question: \emph{What memorization profile do LTMs, inherently trained to perform ICL, have?}    
To investigate memorization capabilities of \tabfms, we present \iclmia and apply MIA. 
Specifically, we test membership of a candidate set of query rows by manipulating $C$ as an active probing mechanism (Fig.~\ref{fig:threat_model}).
We hypothesize that by carefully modifying $C$ to eliminate discriminative information, it disrupts the model's context-derived prediction capabilities.
Forced to abandon in-context reasoning, the model falls back on its memorized parametric knowledge.
For member rows in $D$, we expect predictions to exhibit higher confidence and remain robust to context manipulations \cite{choquette2021label, yeom2018privacy}. 
We identify specific circumstances for memorization in LTMs (Sec.~\ref{sec:probing_memorization}) and discuss protection measures~(Sec.~\ref{sec:conclusion}).

\section{Related Work}

While \tabfms \cite{hollmann2022tabpfn, qu2025tabicl, spinaci2025contexttab, garg2025real} rapidly advance, investigations into their privacy properties remain limited \cite{nayyeri2026survey}. 
Existing \tabfm privacy literature primarily targets test-time evasion and adversarial robustness rather than memorization \cite{simonetto2024constrained, djilani2026robustness, anwar2024adversarial}.
Conversely, ICL privacy research largely focuses on LLMs, probing either the leakage of hidden context examples \cite{wen2024membership, duan2024privacy} or table memorization within text corpora \cite{german2025tabmia, bordt2024elephants}. 
To assess memorization, state-of-the-art MIAs typically rely on shadow models \cite{carlini2022lira, zarifzadeh2024low}. However, training shadow models for large models is computationally prohibitive \cite{hayes2025strong} and yields unreliable false-positive bounds when the pre-training distribution is unknown \cite{zhang2025position}.
Towards privacy assessment of \tabfms, we adapt techniques from robustness-assessing MIA \cite{choquette2021label} and attribute inference (AIA) \cite{jayaraman2022attribute, annamalai2024linear, salem2023sok}. 
While AIA aims to infer an unknown sensitive value given a partial record, our goal is inferring membership of a record. 
To assess memorization in \tabfms, we adapt AIA's multiple-choice testing and label randomization and combine zero-information context manipulations with difficulty calibration via the pre-trained base model \cite{watson2022importance}.

\section{Contributions}\label{sec:contributions}
We provide the first systematic assessment of memorization in \tabfms. Our contributions are:
\begin{itemize}[itemsep=0em, parsep=0em, topsep=-0.5em, partopsep=0em]
    %\item We introduce \iclmia, a framework to probe memorization capabilities of \tabfms, where the user has full control over the context-query pair.
    %%
    %\iclmia performs controlled fine-tuning to establish membership ground truth, and assess memorization via MIA based on a zero-information \emph{multiple-choice protocol} that combines adversarial context manipulations with difficulty calibration.
    %\item We introduce \iclmia, a framework to probe memorization in \tabfms via controlled fine-tuning (for verified membership ground truth) and a zero-information \emph{multiple-choice protocol} combining adversarial context manipulations with difficulty calibration.
    \item We introduce \iclmia, a probing framework designed to separate context-based predictions from parametric memorization. Our zero-information multiple-choice context strips away contextual patterns forcing the model to fall back on parametric memorization.%We perform controlled fine-tuning to establish membership ground truth, and deploy a zero-information context protocol with context manipulations to strip away valid contextual patterns and force the model to fall back on its parametric memory.
    \item We avoid common MIA false positives by accounting for distribution shifts (via data pre-processing), feature contamination (via data deduplication), base-rate confounding (via label randomization), and intrinsic low-complexity samples (via difficulty calibration). 
    %missing difficulty calibration (via pre-trained base model as a reference).
    %We account for false positives in MIA evaluations by avoiding common pitfalls; namely, distribution shifts (via controlled data pre-processing), feature contamination (via data deduplication), base-rate confounding (via structural label randomization), and difficulty calibration (via pre-trained base model as a reference).
    \item We perform an extensive evaluation on \tabfm ConTextTab across 1,128 fine-tuning configurations on 10 diverse classification and regression tasks from CARTE.
    \item We detect moderate memorization in 8 of the 10 tasks (with $\text{AUC} > 0.5$ and $\text{TPR}_{@r\%} > 0.1$ for $r \in \{1, 10\}$) in our controlled setup with single-task fine-tuning and fixed context-query pairs. The signal is higher for low-cardinality and binary tasks with small training query sizes ($Q=50$), but largely vanishes under practical setups (e.g., larger $Q=512$, multi-dataset training, random context-query sampling)~(App.~\ref{app:ablation_realistic}).
    %\tocheck{\item We confirm via a controlled ablation (App.~\ref{app:ablation_realistic}) that memorization degrades systematically as training approaches real-world conditions: joint training on all 10 tasks with random context-query sampling at $Q=512$ reduces detectable tasks from 4 (single-task) to 1, with that sole remaining signal appearing only at Epoch~10 — beyond the 2--5 epoch budget of real \tabfm pre-training~\cite{spinaci2025contexttab}.}
    %\tocheck{This confirms that while \tabfms can memorize facts, appropriate measures can effectively mitigate the risk to sensitive data.}
    %under constructed training setups. Notably, we find that this signal is highest for low-cardinality and binary tasks with small query size in training ($Q=50$), but largely vanishes under realistic training conditions (e.g., larger query sizes at $Q=512$). \tocheck{This confirms that while \tabfms can memorize facts, appropriate measures can effectively mitigate the risk to sensitive data.}
    %\item We identify target cardinality as the primary structural driver of memorization vulnerability: low-cardinality and binary tasks concentrate gradient signal onto individual records and are most susceptible, while high-cardinality tasks resist memorization unless per-step query coverage is sufficiently high.
    %we detect memorization in 22 of the 26 tasks, with detection rates reaching up to 49\% TPR\@10\%FPR. Furthermore, we identify query scaling (e.g., $Q=512$) as a primary structural factor suppressing memorization exposure.
\end{itemize}

Overall, our findings show memorization capabilities in LTMs which help to inform appropriate protection measures.
Next, we introduce our methodology and setup (Sec.~\ref{sec:methodology}), detail our evaluation (Sec.~\ref{sec:eval}), and conclude with limitations, mitigations, and future works (Sec.~\ref{sec:conclusion}). 

\section{Methodology and Setup}\label{sec:methodology}

We introduce \iclmia, a framework designed to probe memorization in \tabfms and mitigate common MIA evaluation pitfalls (Tab.~\ref{tab:pitfalls}, App.~\ref{app:pitfalls}). %Crucially, \iclmia eliminates the need for computationally expensive shadow models by leveraging the pre-trained base model as a zero-shot calibration reference. 
First, we outline unique challenges of assessing \tabfms. Then, we discuss the two stages of \iclmia (Fig.~\ref{fig:methodology}, App.~\ref{app:methodology}). Namely, controlled fine-tuning with careful data pre-processing and fixed query-context pairs across epochs (Sec.~\ref{sec:ground_truth}) and memorization probing with a suite of context manipulations calibrated against the pre-trained base model as reference (Sec.~\ref{sec:probing_memorization}).

%we first describe the specific challenges of applying Membership Inference Attacks (MIAs) to large tabular models. We then introduce \iclmia, a framework designed to probe memorization while systematically addressing common MIA evaluation pitfalls. Specifically, \iclmia avoids the need for computationally expensive shadow models by utilizing the pre-trained base model as a zero-shot calibration reference. 

\para{Challenges in Assessing \tabfms}
%Testing memorization of \tabfms is inherently difficult for two main reasons.
%
%Standard membership inference techniques designed for LLMs fail against \tabfms due to two fundamental barriers. First, \tabfms are explicitly trained to derive predictions from the user-provided context rather than parametric memory. Second, unlike LLMs with massive output vocabularies, \tabfms are encoder-only architectures strictly constrained to outputting probabilities over the specific candidate labels present in the context
%
\tabfms rely on ICL to infer structural feature-target correlations from the context, a mechanism that intuitively acts as a barrier to accessing the model's parametric memory. Furthermore, unlike LLMs with massive output vocabularies, \tabfms are encoder-only architectures strictly constrained to outputting probabilities over the specific candidate labels present in the context.
Our context manipulations address these challenges via zero-information multiple-choice protocol (Sec.~\ref{sec:probing_memorization}). 

%. Unlike LLMs that generate arbitrary text, \tabfms are encoder-only models restricted to outputting probabilities exclusively over the candidate labels from the context.
%
%Second, \tabfms typically heavily preprocess their inputs. For instance, categorical targets are mapped to integers (sorted lexicographically in TabPFN, randomly permuted by frequency rank in ConTextTab), and regression targets are discretized into quantile-based buckets~\cite{hollmann2022tabpfn,spinaci2025contexttab}. As a result, e.g., strings, a typical probing target, cannot be directly memorized.

%\para{The \iclmia Framework}
%To investigate whether table-native ICL models memorize explicit row data or merely learn structural connections, we structure \iclmia into two phases: establishing a memorization ground truth, and probing memorization via the attack phase. The complete pipeline is illustrated in Fig.~\ref{fig:methodology}.

\subsection{Establishing a Memorization Ground Truth}\label{sec:ground_truth}
%
%A general issue for MIAs is the lack of a membership ground truth when the exact training data is unknown \cite{zhang2025position}.
%
%This applies also to \tabfms, where the training data is drawn from large, random subsets of real-world corpora \cite{spinaci2025contexttab}.
%
\iclmia pre-processes data to eliminate artifacts, and performs controlled fine-tuning to induce memorization. 
% membership ground truth.

%To avoid common pitfalls (Tab.~\ref{tab:pitfalls}),  followed by controlled fine-tuning to establish a membership ground truth.
%To establish a ground truth \iclmia

%and avoid common pitfalls (Tab.~\ref{tab:pitfalls}), \iclmia leverages data pre-processing and controlled fine-tuning. 
%
%As shown in Fig.~\ref{fig:methodology}, these steps are followed by a probing phase (Sec.~\ref{sec:probing_memorization}).

%Since the training data of ConTextTab is drawn from random subsets of the T4 dataset \cite{gardner2024large}, the exact records processed during pre-training are unknown \cite{spinaci2025contexttab}. This lack of verified membership ground truth poses a significant challenge for MIA evaluation, as it undermines the ability to reliably bound the false positive rate (FPR) \cite{zhang2025position}.

%\tocheck{The complete pipeline is illustrated in Fig.~\ref{fig:methodology}. This phase covers two steps: \emph{Data Pre-Processing} (to eliminate data artifacts that could confound the membership signal) and \emph{Controlled Fine-Tuning} (to induce memorization under a controlled setup, establishing a verified ground truth). We address the third challenge above by establishing membership ground truth via controlled fine-tuning \cite{evertz2025chasing, zhang2025position}: rather than attacking pre-training data, we fine-tune a dedicated model $M_\theta$ per task so that member and non-member labels are exactly known.}

\para{Data Pre-Processing}
To avoid data artifacts being interpreted as memorization signal, \iclmia carefully pre-processes data.
First, we deduplicate records to prevent \emph{feature contamination}, i.e., identical records appearing in both member and non-member sets, which inflate the memorization signal.
Second, we ensure distributional closeness between disjoint member and non-member splits to preclude \emph{distributional shifts}. Specifically, for target label distributions via total variation distance $\text{TVD} < 0.05$ \cite{gibbs2002choosing} for classification and the Kolmogorov--Smirnov statistic $ \text{KS}<0.05$ \cite{massey1951kolmogorov} for regression (i.e., difference between empirical cumulative density functions) (Tab.~\ref{tab:distribution_stats}, App.~\ref{app:dataset_details}).
%
%Specifically, for target label distributions via total variation distance $\text{TVD} < 0.05$ \cite{gibbs2002choosing} for classification and Kolmogorov-Smirnov (KS) $p > 0.5$ \cite{massey1951kolmogorov} for regression.
%
%We keep only splits with $\text{TVD} < 0.05$ or KS $p$-value $> 0.5$, i.e., rejecting the null hypothesis that the distributions are different.
%
To mitigate the \emph{base-rate fallacy}, i.e., model exploits label distribution to predict likely outcomes, we apply two measures. First, we enforce a uniform label distribution in the training context, preventing the model from minimizing loss by collapsing to majority-class predictions. Second, following \cite{annamalai2024linear}, we apply label randomization, i.e., scrambling target labels while preserving marginal distributions, forcing the model to learn statistically improbable mappings. %Query distributions remain unaltered to prevent further artifacts.

%the target labels of training records are scrambled while maintaining the overall marginal distribution, forcing the model to learn a statistically improbable mapping. For queries, we preserve the source label distribution to avoid introducing additional distributional artifacts.

\para{Controlled Fine-Tuning}
To induce memorization, we fine-tune on fixed context-query pairs for up to 500 epochs.
Specifically, each training step processes a static context set $C'$ with a batch of query rows of size $Q$. 
We fine-tune across varying learning rates $\eta \in \{10^{-1}, 10^{-2}, 10^{-3}, 10^{-4}\}$ and query sizes $Q \in \{50, 512\}$.
We vary $Q$ in training to test its influence on memorization. Smaller $Q$ should amplify the contribution of individual query rows, while larger $Q$ should dilute them. 
Our evaluation supports this dynamic, showing that larger $Q=512$ hinders memorization~(Sec.~\ref{sec:eval}). %, App.~\ref{app:ablation_realistic}).

%This dynamic is also supported by our evaluation results showing that larger $Q=512$ hinders memorization (Sec.~\ref{sec:eval}, App.~\ref{app:ablation_realistic}).
%
%To establish a membership ground truth, we log every exact pair, guaranteeing that only explicitly processed rows are labeled as members. 
%To induce memorization, we fine-tune the model across varying learning rates $\eta \in \{10^{-1}, 10^{-2}, 10^{-3}, 10^{-4}\}$ and query sizes $Q \in \{50, 512\}$.
%
%To boost the memorization signal, we repeatedly expose the model to the same fixed context-query pairs for up to 500 epochs. 
%
%\tocheck{This resolves the third challenge above: since we control the fine-tuning data exactly, membership labels are verified and the FPR can be reliably bounded.}

\subsection{Probing Memorization with \iclmia}\label{sec:probing_memorization}
The probing stage aims to detect memorization signals by combining zero-information context manipulations with difficulty calibration via the base model. 

\para{Multiple-Choice Protocol}
To separate expected context-based model predictions from unintended parametric memorization, we propose a zero-information strategy mimicking a multiple-choice question. 
For each query row $q=(x_q, y_q)$ and candidate label set $Y$, we construct a probing context $C_{base}$ containing exactly $|Y|$ copies of the query's features $x_q$, each paired with a distinct candidate label $y_i \in Y$, thereby providing the model with all possible candidate targets.
Our hypothesis is that if we strip away all valid contextual patterns, the model will be forced to abandon in-context inference and default to its parametric memory, selecting the target value observed during fine-tuning.

\para{Context Manipulations}
To assess robustness of predictions across multiple runs, we subject $C_{base}$ to a suite of 48 context manipulations $\Pi$, including positional biasing, distractor injection, and distributional shifts by, e.g., increasing the counts of true target labels, or non-true labels (detailed in App.~\ref{sec:appendix_manipulations}).
We compute the probe loss $L_k(q) = \mathcal{L}(M_{\theta}(x_q \mid C_k), y_q)$ for each manipulated context $C_k \in \Pi$. The robustness score is defined as the average loss across all manipulations $S(q) = \frac{1}{|\Pi|} \sum_{\pi_k \in \Pi} L_k(q)$, which is the basis for our memorization metric. 
% 
%These metrics remain robust discriminators even when raw loss inverts.

\para{Difficulty Calibration}
An inherently easy query (e.g., with obvious target from its features) has a low loss, regardless of its membership in $D$. To account for this, we follow the \emph{difficulty calibration} approach of Watson et al.~\cite{watson2022importance} and compare robustness score from the fine-tuned model, $S(q)$, to that of the pre-trained base model, $S_{ref}(q)$, and compute its delta as $\Delta S(q) = S(q) - S_{ref}(q)$. 
Notably, we compute $S(q)$ not only using the loss but also using distributional metrics, i.e., prediction entropy and target confidence, which remain robust discriminators even with fine-tuning-induced artifacts (e.g., loss inversion, Sec.~\ref{sec:eval}). 
%Crucially, since fine-tuning can induce data artifacts like loss inversion (where member loss paradoxically exceeds non-member loss, Sec.~\ref{sec:eval}), we compute our robustness score $S(q)$ using distributional metrics, i.e., prediction entropy and target confidence.

\para{Memorization Metric}
We compute AUC over the $\Delta S$ scores (denoted $\text{AUC}(\Delta S)$ or simply $\text{AUC}$), and consider $\text{AUC} > 0.5$ as the threshold for \emph{detectable} memorization.
Unlike high AUC baselines for LLMs \cite{duan2024privacy}, \tabfms operate over a constrained output space. Hence, we adopt this threshold to capture any calibrated signal above chance \cite{liu2025urania}. 
Furthermore, since global AUC averages across all records and can obscure localized memorization \cite{carlini2022lira}, we report the true positive rate at low false positive rates ($\text{TPR}_{@r\%}$ for $r \in \{1, 10\}$) providing a granular view of more confident signals. %the critical low-FPR regime.
\section{Empirical Evaluation}\label{sec:eval}

Next, we summarize our results of \iclmia, and defer full details to App.~\ref{app:exhaustive_results} due to space constraints.
%Due to space constraints we defer full details to App.~\ref{app:exhaustive_results}.
%Next, we analyze the results of our preliminary evaluation of \iclmia. 
%
%First, we outline the experimental setup.
%
%Then, we isolate memorization outcomes and analyze the configurations that yield memorization signal shown in Tab.~\ref{tab:main_results} (with full results in App.~\ref{app:exhaustive_results}).

%and categorize the possible evaluation outcomes.
%
%Then, we isolate outcomes driven by fine-tuning artifacts 
%(i.e., representational collapse and loss inversion at extreme learning rates), 
%and analyze the configurations that yield memorization signal.

\para{Setup}
We evaluate \iclmia on ConTextTab \cite{spinaci2025contexttab} on 10 tasks from CARTE \cite{kim2024carte}. We select these tasks as they were not used in pre-training and represent a wide variety of domains and data distributions (details in Tab.~\ref{tab:datasets}).
We test for memorization across epochs to capture the evolution of the memorization over time resulting in 1128 probing runs (configurations in Tab.~\ref{tab:hyperparameters}). % across all configurations (Tab.~\ref{tab:hyperparameters}).

\para{Isolating Memorization Outcomes}
Our evaluation yields five distinct outcomes (summarized in Tab.~\ref{tab:attack_classification}).
{%
\begin{table}[t]
\centering
\caption{ \iclmia detectable memorization configurations (Q=50 except $^\dagger$ with $Q=512$). Full results in Tabs.~\ref{tab:appendix_exhaustive_q50}--\ref{tab:appendix_exhaustive_q512} in
App.~\ref{app:exhaustive_results}.} % (\textbf{bold} = highest signal).\todo{in caption can refer to Tab 9 or App for full details (want to avoid that reviewers miss that this is just an excerpt)}}
\vspace{-0.5em}
\label{tab:main_results}
\scriptsize
\setlength{\tabcolsep}{0.35em}
\begin{tabular}{llc r r r}
\textbf{Dataset} & \textbf{LR} & \textbf{Ep.}
  %& $\Delta\mathcal{L}_v^M$
  & $\text{AUC}(\Delta S)$
  & $\text{TPR}_{@10\%}$
  & $\text{TPR}_{@1\%}$ \\
\midrule
%\texttt{coffee\_ratings}          
%& $10^{-3}$ 
%& 10 
%%& $0.18$     
%& $0.56$ 
%& $0.20$ 
%& $0.00$ \\
\texttt{buy\_buy\_baby}           
& $10^{-4}$ 
& 10 
%& ${-}0.06$  
& $0.61$ 
& $0.14$ 
& $0.12$ \\
\texttt{chocolate\_bar}           
& $10^{-3}$ 
& 10 
%& $0.02$     
& $0.61$ 
& $0.18$ 
& $0.02$ \\
\texttt{babies\_r\_us}            
& $10^{-4}$ 
& 50 
%& ${-}0.18$  
& $\mathbf{0.66}$
& $0.02$ 
& $0.00$ \\
%\midrule
\texttt{beer\_ratings}$^\dagger$  
& $10^{-4}$ 
& 50 
%& $0.24$     
& $0.59$ 
& $\mathbf{0.49}$ 
& $\mathbf{0.17}$ \\
\texttt{bikedekho}$^\dagger$      
& $10^{-4}$ 
& 10 
%& ${-}0.02$  
& $0.58$ 
& $0.15$ 
& $0.02$ \\
\end{tabular}
\vspace{-2.0em}
\end{table}%
}

Impractical fine-tuning configurations (e.g., large learning rate) can induce data artifacts mimicking memorization. The most prominent is \emph{representational collapse} (affecting ${\approx}55\%$ of runs at $Q=50$), where aggressive learning rates ($\lr \ge 10^{-2}$) destroy predictive capabilities, yielding near-constant outputs that artificially inflate loss. 
Difficulty calibration identifies these false positives: if $M_{\text{ref}}$ achieves identical zero-shot separation, the signal reflects pre-existing dataset bias, not memorization.
%
%We identify these false positives via difficulty calibration: if the pre-trained base model $M_{\text{ref}}$ achieves the same member/non-member separation without ever seeing the fine-tuning data, the signal is a pre-existing dataset bias, not parametric memorization. 
%
A secondary artifact is \emph{loss inversion} (${\approx}19\%$ of runs), where pre-trained structural biases drive member loss higher than non-member loss. %, which we mitigate using robust distributional metrics.
A detailed breakdown of all outcomes, including context overfitting, is deferred to App.~\ref{app:secondary_dynamics}, with results for all datasets in Tabs.~\ref{tab:appendix_exhaustive_q50}--\ref{tab:appendix_exhaustive_q512}.
After accounting for these fine-tuning artifacts, we isolate memorization signal in \enx{12.0\%} of runs at $Q=50$ (Tab.~\ref{fig:aggregate_outcomes}), and 8 of the 10 evaluated tasks in total under typical fine-tuning configurations ($\lr \le 10^{-3}$), exemplified in Tab.~\ref{tab:main_results}.
More granularly, the TPR at low FPR metric reveals localized memorization. 
On \texttt{beer\_ratings} at $Q=512$, \iclmia achieves only $\text{AUC} = 0.59$ but identifies 49\% of member records at 10\% FPR; this dataset retains detectable memorization at $Q=512$ due to its high target cardinality and small size, as analyzed below. On \texttt{buy\_buy\_baby}, we reach $\text{TPR}_{@1\%} = 0.12$, indicating that in some instances fine-tuning records can be confidently distinguished.

%that fine-tuning records can be distinguished with high precision at low FPR. 
%probe loss is surprisingly higher for members than non-members after fine-tuning, driven by pre-trained structural biases, which we mitigate by using distributional metrics.
%
%We address this by using distributional metrics (i.e., prediction entropy and confidence) which remain robust discriminators even when loss is inverted.
%

\para{Drivers of Memorization}
We find that memorization depends on the interaction between query size $Q$, target cardinality $k$, and dataset size $n$. 
Firstly, expanding $Q$ mitigates memorization in our evaluation: at $Q=512$, the overall detectable rate drops to \enx{1.8\%} (a $6.7\times$ reduction versus $Q=50$), as larger queries stabilize batch variance and enforce stronger zero-shot inductive biases.
Secondly, at low query sizes ($Q=50$), memorization is driven by \emph{label density} ($n/k$, the average records per class). Binary tasks ($k=2$, e.g., \texttt{chocolate\_bar\_ratings}) concentrate the gradient signal for specific classes per step, yielding the highest vulnerability (up to \enx{55\%} detectable runs). As $k$ grows, this concentration dissolves.
Finally, high-cardinality tasks require two conditions to trigger memorization: high \emph{query coverage} ($Q/n$, the dataset fraction seen per step) and a target that is \emph{not structurally deducible}. If a target cannot be logically inferred from its features alone (e.g., exact prices or subjective float ratings in \texttt{beer\_ratings}), the model is forced to fall back on parametric memory, provided $Q$ covers enough of the dataset. Conversely, if a target can be deduced via ICL (e.g., text-statistic formulas in \texttt{clear\_corpus}), the model relies on in-context reasoning and yields no memorization signal, regardless of~$Q$.

\begin{figure}[t]
\centering
\scalebox{0.87}{% adjust to fit available space
\begin{tikzpicture}
    \begin{axis}[
        width=0.85\columnwidth,
        height=13.0em,
        xlabel={Fine-Tuning Epochs},
        ylabel={\% of Configurations},
        xmode=log,
        ymode=log,
        log ticks with fixed point,
        xmin=1, xmax=500,
        ymin=0.5, ymax=100,
        xtick={1, 2, 10, 50, 100, 250, 500},
        ytick={1, 2, 5, 10, 20, 50},
        legend style={
            at={(1.02,1)}, anchor=north west,
            font=\scriptsize, draw=none,
            legend cell align=left,
        },
        grid=major,
        grid style={dashed, gray!30},
        thick
    ]
    % Q=50 (solid lines)
    \addplot[color=gray!80!black, mark=square*, mark options={scale=0.8}] coordinates {(1,53.3) (2,61.3) (10,59.2) (50,56.7) (100,51.2) (250,40.0)};
    \addlegendentry{Collapse}

    \addplot[color=orange, mark=triangle*, mark options={scale=1.1}] coordinates {(1,20.8) (2,20.0) (10,15.0) (50,19.2) (100,20.0) (250,25.0)};
    \addlegendentry{Inversion}

    \addplot[color=red, mark=*, line width=1.5pt] coordinates {(1,4.2) (2,3.8) (10,15.0) (50,15.8) (100,17.5) (250,20.0)};
    \addlegendentry{Det. Mem.}

    \addplot[color=blue, mark=diamond*, mark options={scale=1}] coordinates {(1,6.7) (2,10.0) (10,5.0) (50,3.3) (100,3.8) (250,5.0)};
    \addlegendentry{Ctx.\ Ovfit.}

    \addplot[color=teal, mark=o] coordinates {(1,15.0) (2,5.0) (10,5.8) (50,5.0) (100,7.5) (250,10.0)};
    \addlegendentry{No Signal}

    % Q=512 (dashed lines, same colors)
    \addplot[forget plot, dashed, color=gray!80!black, mark=none] coordinates {(1,53.6) (2,55.1) (10,55.0) (50,53.6) (100,47.1) (250,45.7) (500,26.7)};

    \addplot[forget plot, dashed, color=orange, mark=none] coordinates {(1,21.8) (2,18.8) (10,15.6) (50,16.4) (100,22.9) (250,22.9) (500,30.0)};

    \addplot[forget plot, dashed, color=red, mark=none, line width=1.5pt] coordinates {(1,1.8) (10,1.8) (50,5.5) (100,0.0) (250,0.0) (500,0.0)};

    \addplot[forget plot, dashed, color=blue, mark=none] coordinates {(1,0.9) (2,2.9) (10,9.2) (50,7.3) (100,10.0) (250,10.0) (500,10.0)};

    \addplot[forget plot, dashed, color=teal, mark=none] coordinates {(1,21.8) (2,23.2) (10,18.3) (50,17.3) (100,20.0) (250,21.4) (500,33.3)};

    % Line style legend
    \addlegendimage{solid, gray, thick} \addlegendentry{$Q=50$}
    \addlegendimage{dashed, gray, thick} \addlegendentry{$Q=512$}
    \end{axis}
\end{tikzpicture}%
}% end \scalebox
\vspace{-0.5em}
\caption{Temporal evolution of \iclmia outcomes} % across the 10 CARTE tasks.} % $Q=512$ detectable memorization line ends at epoch~50 (no detectable signal beyond that epoch).}
\label{fig:epoch_evolution}
\vspace{-1.5em}
\end{figure}
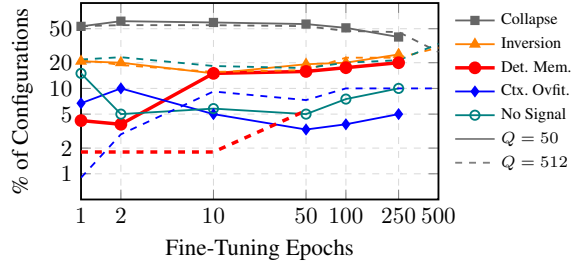

\para{Temporal Evolution of Memorization}
Across fine-tuning epochs (Fig.~\ref{fig:epoch_evolution}), the $Q=50$ signal emerges at epoch 1 (\enx{4.2\%} of configurations) and grows to \enx{20\%} by epoch 250. At $Q=512$, memorization is largely suppressed: it peaks at \enx{5.5\%} around epoch 50 and vanishes after, showing that larger queries reduce memorization and accelerate its decay. 

\section{Discussion and Conclusion}\label{sec:conclusion}

%In this section, we discuss limitations of \iclmia, mitigations, and directions for future work.
Next, we discuss limitations, mitigations, and future work.

\para{Limitations}
First, while \iclmia detects memorization, the signal is moderate: only 20\% of tasks show $\text{TPR}_{@1\%} > 0.1$. % \cite{carlini2022lira}. 
Second, our probe uses an artificial setup (single task, fixed context-query pairs) to induce memorization. 
In contrast, realistic \tabfm regimes pre-train on large task mixtures for only 2--5 epochs \cite{spinaci2025contexttab}, which dilutes memorization. 
We empirically verify that under joint multi-task training with random context-query sampling at $Q=512$, detectable tasks drop from 5 to 1, with the signal appearing only at epoch~10, i.e., beyond realistic budgets (App.~\ref{app:ablation_realistic}). Finally, while our evaluation is broad, our scope is currently limited (1 model, 10 tasks) and to be expanded.

\para{Mitigations}
To protect sensitive data, appropriate measures must be taken.
Differential privacy (e.g., DP-SGD~\cite{abadi2016deep}) provides rigorous privacy guarantees that  requires careful privacy-utility trade-offs but also shows promising utility when fine-tuning a public base model \cite{yu2021differentially, li2021large}.
Additionally, our findings show \tabfm memorization in certain training conditions that can be avoided (e.g., small query size with fixed query-context pairs over many epochs on single task, App.~\ref{app:ablation_realistic}) or inform data filtering of memorization-prone samples (albeit with potential for privacy onion effects \cite{carlini2022privacy}).

\para{Conclusion and Future Work}
We initiated the systematic study of memorization in \tabfms. In our controlled setup \iclmia detects moderate memorization signal in 8 out of 10 evaluated tasks ($\text{AUC}$ up to  $0.67$ and TPR at $1\%$ FPR $>0.1$).
Our results demonstrate that, under specific conditions, tabular foundation models can memorize data, underscoring the need for appropriate protective measures when sensitive data is involved.
In future work, we aim to expand the evaluation to other LTMs \cite{qu2025tabicl,hollmann2022tabpfn} and datasets \cite{klein2024salt}.  
%, and assess whether dynamic context sampling alleviates residual memorization for high-cardinality tasks. 
We also aim to design \emph{data canaries} (i.e., crafted samples with unique fingerprints or improbable feature-label mappings \cite{carlini2019secret}) to amplify memorization signal and bound the worst-case detection limits of \iclmia.

\bibliography{bibliography}
\bibliographystyle{icml2026}

%%%%%%%%%%%%%%%%%%%%%%%%%%%%%%%%%%%%%%%%%%%%%%%%%%%%%%%%%%%%%%%%%%%%%%%%%%%%%%%
%%%%%%%%%%%%%%%%%%%%%%%%%%%%%%%%%%%%%%%%%%%%%%%%%%%%%%%%%%%%%%%%%%%%%%%%%%%%%%%
% APPENDIX
%%%%%%%%%%%%%%%%%%%%%%%%%%%%%%%%%%%%%%%%%%%%%%%%%%%%%%%%%%%%%%%%%%%%%%%%%%%%%%%
%%%%%%%%%%%%%%%%%%%%%%%%%%%%%%%%%%%%%%%%%%%%%%%%%%%%%%%%%%%%%%%%%%%%%%%%%%%%%%%
\newpage
\appendix
\onecolumn

\section{Detailed Probing Methodology and Pipeline}\label{app:methodology}

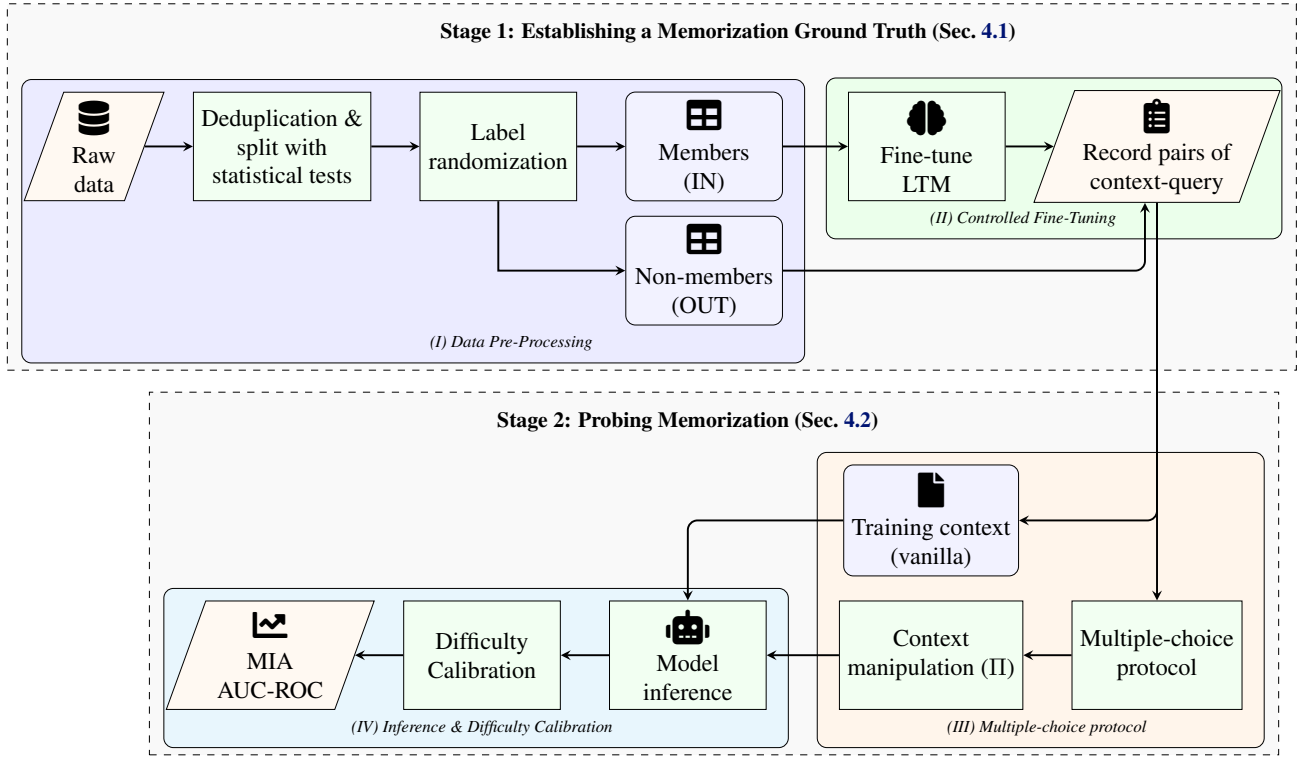
\begin{figure*}[htbp]
\centering
\resizebox{\textwidth}{!}{%
\begin{tikzpicture}[
  node distance=2em and 2em,
  box/.style={rectangle, draw, rounded corners, minimum width=6.5em, minimum height=4.5em, align=center, fill=blue!5},
  process/.style={rectangle, draw, minimum width=6.5em, minimum height=4.5em, align=center, fill=green!5},
  io/.style={trapezium, trapezium left angle=70, trapezium right angle=110, draw, minimum width=5em, minimum height=4.5em, align=center, fill=orange!5},
  arrow/.style={thick, ->, >=stealth}
]

% Phase 1 Nodes
\node[io] (data) {\Large\faDatabase\\[0.1em]Raw\\data};
\node[process, right=of data] (split) {Deduplication \& \\split with \\statistical tests};
\node[process, right=of split] (label_rand) {Label\\randomization};
\node[box, right=of label_rand] (members) {\Large\faTable\\[0.1em]Members\\(IN)};
\node[process, right=of members, xshift=0.7em] (finetune) {\Large\faBrain\\[0.1em]Fine-tune\\LTM};
\node[io, right=of finetune] (logger) {\Large\faClipboardList\\[0.1em]Record pairs of\\ context-query};

\node[box, below=0.6em of members] (nonmembers) {\Large\faTable\\[0.1em]Non-members\\(OUT)};

% Phase 1 Arrows
\draw[arrow] (data) -- (split);
\draw[arrow] (split) -- (label_rand);
\draw[arrow] (label_rand) -- (members);
\draw[arrow] (label_rand) |- (nonmembers);
\draw[arrow] (members) -- (finetune);
\draw[arrow] (finetune) -- (logger);
%\draw[arrow, rounded corners] (members.north) -- ++(0, 1.5em) -| ([xshift=-1.5em]logger.north);
\draw[arrow, rounded corners] (nonmembers.east) -| ([xshift=-0.5em]logger.south);

% Phase 2 Nodes (Right to Left)
\node[process, below=16.5em of logger] (protocol) {Multiple-choice\\protocol};
\node[process, left=of protocol] (perturb) {Context\\manipulation ($\Pi$)};
\node[process, left=3em of perturb] (inference) {\Large\faRobot\\[0.1em]Model\\inference};
\node[process, left=of inference] (aggregate) {Difficulty\\Calibration};
\node[io, left=of aggregate] (auc) {\Large\faChartLine\\[0.1em]MIA\\AUC-ROC};

% New Vanilla Context Node
\node[box, above=1em of perturb] (vanilla) {\Large\faFile\\[0.1em]Training context\\(vanilla)};

% Phase 2 Arrows
\draw[arrow] (logger) -- (protocol);
\draw[arrow] (protocol) -- (perturb);
\draw[arrow] (perturb) -- (inference);
\draw[arrow] (inference) -- (aggregate);
\draw[arrow] (aggregate) -- (auc);

% Arrows for Vanilla Context Path
\draw[arrow, rounded corners] (logger.south) |- (vanilla.east);
\draw[arrow, rounded corners] (vanilla.west) -| (inference.north);

% Phase and sub-phase labels (detached for independent positioning)
\node[font=\scriptsize\itshape, below=0.4em of nonmembers, xshift=-8.0em, inner sep=0.0em] (lbl_sub1) {(I) Data Pre-Processing};
\node[font=\scriptsize\itshape, below=0.4em of finetune, inner sep=0.0em, xshift=4.0em] (lbl_sub2) {(II) Controlled Fine-Tuning};
\node[font=\scriptsize\itshape, below=0.4em of perturb, xshift=4.8em, inner sep=0.0em] (lbl_sub3) {(III) Multiple-choice protocol};
\node[font=\scriptsize\itshape, below=0.4em of aggregate, inner sep=0.0em] (lbl_sub4) {(IV) Inference \& Difficulty Calibration};
\node[font=\small\bfseries, above=2.0em of members, xshift=1.0em, inner sep=0.0em] (lbl_phase1) {\textbf{Stage 1: Establishing a Memorization Ground Truth} (Sec.~\ref{sec:ground_truth})};
\node[font=\small\bfseries, above=7.0em of inference, inner sep=0.0em] (lbl_phase2) {\textbf{Stage 2: Probing Memorization} (Sec.~\ref{sec:probing_memorization})};

% Backgrounds
\begin{scope}[on background layer]
  % Outer phase boxes
  \node[draw, dashed, inner ysep=.8em, inner xsep=1.5em, fit=(data) (split) (members) (finetune) (logger) (nonmembers) (lbl_sub1) (lbl_sub2) (lbl_phase1), fill=gray!5] (phase1) {};
  \node[draw, dashed, inner ysep=.8em, inner xsep=1.5em, fit=(protocol) (perturb) (inference) (aggregate) (auc) (vanilla) (lbl_sub3) (lbl_sub4) (lbl_phase2), fill=gray!5] (phase2) {};
  % Phase 1 subphases
  \node[draw, rounded corners, inner ysep=0.5em, inner xsep=0.9em, fit=(data) (split) (members) (nonmembers) (lbl_sub1), fill=blue!8] {};
  \node[draw, rounded corners, inner ysep=0.5em, inner xsep=0.9em, fit=(finetune) (logger) (lbl_sub2), fill=green!8] {};
  % Phase 2 subphases
  \node[draw, rounded corners, inner ysep=0.5em, inner xsep=0.9em, fit=(protocol) (perturb) (vanilla) (lbl_sub3), fill=orange!8] {};
  \node[draw, rounded corners, inner ysep=0.5em, inner xsep=0.9em, fit=(inference) (aggregate) (auc) (lbl_sub4), fill=cyan!8] {};
\end{scope}
\end{tikzpicture}
}
\caption{End-to-end \iclmia pipeline. \textbf{Stage 1}: (I)~\emph{Data pre-processing} — deduplication, statistical split validation, member/non-member assignment, and optional label randomization as a null-hypothesis control (applicable to members, non-members, or both); (II)~\emph{Controlled fine-tuning} — LTM fine-tuning on static context-query pairs and logging of all context-query pairs for verified ground truth. \textbf{Stage 2}: (III)~\emph{Multiple-choice protocol} — zero-information context construction and adversarial context manipulation suite $\Pi$; (IV)~\emph{Inference \& difficulty calibration} — model inference under both the training (vanilla) context and $\Pi$, difficulty calibration with instance-level delta score $\Delta S(q) = S(q) - S_{ref}(q)$ computation against the pre-trained reference model $M_{ref}$.}
\label{fig:methodology}
\end{figure*}
Figure~\ref{fig:methodology} illustrates the \iclmia pipeline, which consists of two main stages.
(I) Establishing a memorization ground truth via data pre-processing and controlled fine-tuning. Here, we create a verified member/non-member split, optionally randomize labels as a null-hypothesis control, and log all context-query pairs for ground-truth verification.
(II) Probing memorization using a multiple-choice protocol with adversarial context manipulation. Here, we build a zero-information context, apply a suite of 48 context manipulations, perform model inference under both the vanilla and manipulated contexts, aggregate scores, and compute delta scores against a pre-trained reference model for difficulty calibration.    
%dataset processing, context manipulation, and score computation. 
%The pipeline is designed to systematically evaluate the presence of memorization signals in \tabfms under adversarial context manipulation.
%
\begin{table*}[htbp]
\centering
\caption{Experimental hyperparameters for LTM fine-tuning and the \iclmia protocol.}
\label{tab:hyperparameters}
\begin{tabular}{@{}lll@{}}
\textbf{Stage} & \textbf{Parameter} & \textbf{Value(s)} \\
\midrule
\multirow{9}{*}{\textbf{Fine-Tuning}} & Base Model & \texttt{ConTextTab} \\
& Optimizer & AdamW \\
& Learning Rates ($\lr$) & $\{10^{-1}, 10^{-2}, 10^{-3}, 10^{-4}\}$ \\
& Maximum Epochs & 500 \\
& Max context size & 512 rows \\
& Context label distribution & Uniform \\
& Query size ($Q$) & $\{50, 512\}$ rows \\
& Query label distribution & Source distribution \\
& Member label randomization & \{False, True\} \\
\midrule
\multirow{5}{*}{\textbf{Probing}} & Reference model ($M_{ref}$) & Pre-trained \texttt{ConTextTab} \\
& Epochs used for probing & \{0, 1, 2, 5, 10, 50, 100, 250, 500\} \\
& Context size for probing & Depends on the number of unique labels and adversarial manipulations \\
& Query size for probing & 1 query row replicated 50 times \\
& Adversarial manipulations ($|\Pi|$) & 48 distinct configurations (see Tab.~\ref{tab:perturbations}) \\
& Label randomization & \{False, True, Only Members\} \\
& Evaluation metrics & Cross-entropy loss, entropy, confidence, $R^2$ \\
& Primary summary metric & $\text{AUC}(\Delta S)$, where $\Delta S(q) = S(q) - S_{ref}(q)$ \\
\end{tabular}
\end{table*}

Tab.~\ref{tab:hyperparameters} lists all fine-tuning and probing hyperparameters used in our evaluation.
Tab.~\ref{tab:attack_classification} defines the classification criteria for the five evaluation outcomes. %, applied in priority order.

\begin{table*}[htbp]
\centering
\footnotesize
\caption{Classification criteria for the five evaluation outcome categories, applied in priority order (top to bottom). $\text{AUC}(\Delta S)$ is computed on per-instance delta scores $\Delta S(q) = S(q) - S_{ref}(q)$; $\mathcal{L}_v(M, q)$ denotes the \emph{vanilla loss} (loss on the unmanipulated context, without any $\pi_k$); $\mu_{\text{gap}}$ and $\sigma_{\text{gap}}$ are the mean and standard deviation of the per-manipulation probe loss gap $g_k = \bar{\ell}^{M}_{\text{atk}}(M_\theta, \pi_k) - \bar{\ell}^{NM}_{\text{atk}}(M_\theta, \pi_k)$ across all $\pi_k \in \Pi$.}
\label{tab:attack_classification}
\begin{tabular}{@{}lp{15em}p{26em}@{}}
\textbf{Outcome} & \textbf{Key Condition} & \textbf{Intuition} \\
\midrule
Representational Collapse
  & $\mathcal{L}_v(M_\theta) > 5 \times \mathcal{L}_v(M_{ref})$, or accuracy collapses
  & Fine-tuning destroys predictive ability; the model outputs near-constant predictions. Any separation is a numerical artifact, not memorization. Signal persists unchanged when labels are shuffled. \\
\addlinespace
Loss Inversion
  & $\bar{\ell}^{NM}_{\text{atk}}(M_\theta) < \bar{\ell}^{M}_{\text{atk}}(M_\theta)$
  & A pre-trained structural bias causes non-members to have lower mean probe loss than members across all context manipulations. Loss-based MIA signals are unreliable; distributional metrics (e.g., entropy) should be checked instead. \\
\addlinespace
Detectable Memorization
  & $\text{AUC}(\Delta S) > 0.5$ on loss or entropy
  & The fine-tuned model exposes a calibrated membership signal above the pre-trained baseline. The signal weakens materially under shuffled labels, confirming it tracks actual label content. \\
\addlinespace
Context Overfitting
  & $|\mu_{\text{gap}}| < 0.05$ and $\sigma_{\text{gap}} < 0.2$
  & Fine-tuning encodes dataset schema or label distribution, not individual records. Every context manipulation yields the same near-zero member--non-member probe loss gap (stable null). \\
\addlinespace
No Signal
  & Catch-all (none of the above)
  & $\text{AUC}(\Delta S) \le 0.5$ on all metrics with high variance across manipulations ($\sigma_{\text{gap}}$ large). The model does not systematically distinguish members from non-members under any context manipulation. \\
\end{tabular}
\end{table*}

\subsection{Addressing Common Pitfalls}\label{app:pitfalls}
Tab.~\ref{tab:pitfalls} summarizes the four common MIA evaluation pitfalls identified by \cite{evertz2025chasing} and the corresponding countermeasures implemented in \iclmia.

\begin{table*}[t]
\centering
\footnotesize
\caption{Common MIA evaluation pitfalls from \cite{evertz2025chasing, zhang2025position} and \iclmia corresponding countermeasures.}
\label{tab:pitfalls}
\begin{tabular}{@{}lp{15em}p{26em}@{}}

\textbf{Pitfall} & \textbf{Issue} & \textbf{Countermeasure in \iclmia} \\
\midrule
Base-rate fallacy
  & Majority-class prediction is mistaken for a membership signal.
  & Label randomization forces the model to learn improbable mappings; detectable memorization is the only way to predict the randomized label. \\
\addlinespace
Distributional shift
  & Train/test distributions differ, making the attack detect the shift rather than memorization.
  & Statistical split validation via TVD (classification) and the KS test (regression) ensures members and non-members are identically distributed. \\
\addlinespace
Feature contamination
  & Duplicate rows inflate the attack score by appearing in both member and non-member sets.
  & Strict deduplication is enforced before partitioning. \\
\addlinespace
No calibration
  & Low sample complexity (easy queries) is confused with memorization.
  & Adapting the difficulty calibration idea of \cite{watson2022importance} to our shadow-model-free setting, we compute per-instance delta scores $\Delta S(q) = S(q) - S_{ref}(q)$ using the pre-trained reference model $M_{ref}$; only $\text{AUC}(\Delta S) > 0.5$ is treated as a detectable signal. \\
\addlinespace
Unverified ground truth
  & The exact set of records processed during training is unknown, making it impossible to bound the FPR without shadow models \cite{zhang2025position}.
  & We log every context-query index pair processed during fine-tuning. Only rows provably seen by the model are labeled as members, eliminating label noise from random subsampling or on-the-fly data filtering. \\
\end{tabular}
\end{table*}

\subsection{Dataset Details}\label{app:dataset_details}
Table~\ref{tab:datasets} lists the 10 tasks evaluated in our experiments, all drawn from the CARTE benchmark \cite{kim2024carte}. We selected these datasets to cover a heterogeneous mix of real-world classification and regression domains. Crucially, all evaluated datasets are strictly excluded from the ConTextTab pre-training corpus to ensure a valid zero-shot baseline.

\begin{table}[htbp]
\centering
\caption{Evaluated datasets with target column and task type, i.e., C(lassification) and R(egression), from CARTE \cite{kim2024carte}}
\label{tab:datasets}
\begin{tabular}{@{}lll@{}}
\textbf{Table Name} & \textbf{Target} & \textbf{Type} \\
\midrule
\texttt{anime\_planet} & \texttt{Rating\_Score} & R \\
\texttt{babies\_r\_us} & \texttt{price} & R \\
\texttt{beer\_ratings} & \texttt{review\_overall} & R \\
\texttt{bikedekho} & \texttt{price} & R \\
\texttt{bikewale} & \texttt{price} & R \\
\texttt{buy\_buy\_baby} & \texttt{price} & R \\
\texttt{cardekho} & \texttt{price} & R \\
\texttt{chocolate\_bar\_ratings} & \texttt{Rating} & C \\
\texttt{clear\_corpus} & \texttt{BT\_Easiness} & R \\
\texttt{coffee\_ratings} & \texttt{rating} & C \\
\end{tabular}%
\end{table}

\para{Member/Non-Member Split Validation}
Tab.~\ref{tab:distribution_stats} reports distributional similarity between the member and non-member splits for all 10 datasets at both $Q=50$ and $Q=512$.
For regression, we calculate the Kolmogorov-Smirnov (KS) statistic \cite{massey1951kolmogorov} on the continuous target values, i.e., the maximum difference between the empirical cumulative distribution functions, to bound the distance between splits.
For classification, we compute the Total Variation Distance (TVD) \cite{gibbs2002choosing} over the label distribution.
All KS statistics are below $0.05$ (maximum: $0.035$ for \texttt{buy\_buy\_baby} at $Q=512$).
Furthermore, all KS $p$-values are above the $0.05$ significance threshold (minimum: $0.530$ for \texttt{buy\_buy\_baby} at $Q=512$), meaning we fail to reject the null hypothesis that the member and non-member splits are drawn from the same distribution. This confirms that no statistically significant distributional shift exists between our data partitions. 
%All KS $p$-values are above the $0.05$ significance threshold (minimum: $0.530$ for \texttt{buy\_buy\_baby} at $Q=512$) and the KS statistics are $<0.05$ (maximum: $0.035$ for \texttt{buy\_buy\_baby} at $Q=512$), confirming that no statistically significant distributional shift exists between member and non-member splits.
%
Similarly, TVD values for classification tasks are small ($\leq 0.056$), indicating near-identical label distributions across splits.
We additionally report the Wasserstein distance for completeness \cite{marek2025benchmarking}; note that this metric is scale-dependent and therefore not directly comparable across datasets. % with different target ranges.

\begin{table}[t]
\centering
\caption{Member vs.\ non-member distributional similarity per dataset and query size. Regression tasks use a two-sample KS test; classification tasks use TVD over the label distribution. All KS stat values are $<0.05$ and all KS $p$-values are $>0.05$, confirming no statistically significant distributional shift between member and non-member splits. TVD values for classification tasks are similarly small ($\leq 0.056$), indicating near-identical label distributions across splits. We additionally report the Wasserstein distance for completeness, but note that this metric is scale-dependent and not directly comparable across datasets with different target ranges.}
\label{tab:distribution_stats}
\scriptsize
\setlength{\tabcolsep}{0.4em}
\begin{tabular}{l l r r r r r}
\textbf{Dataset} & \textbf{Task} & $Q$ & \textbf{KS stat} & \textbf{KS $p$} & \textbf{W} & \textbf{TVD} \\
\midrule
\multicolumn{7}{c}{\textit{$Q = 512$}} \\
\midrule
\texttt{anime\_planet}          & regr. & 512 & 0.029 & 0.611 & 0.025 & ---  \\
\texttt{babies\_r\_us}          & regr. & 512 & 0.021 & 0.968 & 0.030 & ---  \\
\texttt{beer\_ratings}          & regr. & 512 & 0.021 & 0.975 & 0.021 & ---  \\
\texttt{bikedekho}              & regr. & 512 & 0.025 & 0.889 & 0.014 & ---  \\
\texttt{bikewale}               & regr. & 512 & 0.033 & 0.595 & 0.011 & ---  \\
\texttt{buy\_buy\_baby}         & regr. & 512 & 0.035 & 0.530 & 0.065 & ---  \\
\texttt{cardekho}               & regr. & 512 & 0.028 & 0.763 & 0.007 & ---  \\
\texttt{clear\_corpus}          & regr. & 512 & 0.023 & 0.941 & 0.032 & ---  \\
\texttt{chocolate\_bar\_ratings}& class. & 512 & ---   & ---   & ---   & 0.044 \\
\texttt{coffee\_ratings}        & class. & 512 & ---   & ---   & ---   & 0.056 \\
\midrule
\multicolumn{7}{c}{\textit{$Q = 50$}} \\
\midrule
\texttt{anime\_planet}          & regr. &  50 & 0.026 & 0.983 & 0.024 & ---  \\
\texttt{babies\_r\_us}          & regr. &  50 & 0.020 & 1.000 & 0.023 & ---  \\
\texttt{beer\_ratings}          & regr. &  50 & 0.025 & 0.993 & 0.016 & ---  \\
\texttt{bikedekho}              & regr. &  50 & 0.027 & 0.983 & 0.012 & ---  \\
\texttt{bikewale}               & regr. &  50 & 0.044 & 0.622 & 0.013 & ---  \\
\texttt{buy\_buy\_baby}         & regr. &  50 & 0.027 & 0.984 & 0.041 & ---  \\
\texttt{cardekho}               & regr. &  50 & 0.032 & 0.921 & 0.010 & ---  \\
\texttt{clear\_corpus}          & regr. &  50 & 0.027 & 0.988 & 0.038 & ---  \\
\texttt{chocolate\_bar\_ratings}& class. &  50 & ---   & ---   & ---   & 0.005 \\
\texttt{coffee\_ratings}        & class. &  50 & ---   & ---   & ---   & 0.043 \\
\end{tabular}
\end{table}

\subsection{Context Manipulations}\label{sec:appendix_manipulations}
The probing stage subjects each multiple-choice context to a structured suite of 48 manipulations $\Pi$, designed to degrade context utility and expose any reliance on parametric memory. Tab.~\ref{tab:perturbations} enumerates all configurations across 11 categories of manipulation types.
Below we describe the 4 main categories.
\begin{enumerate}
    \item \textbf{Positional Shuffling:} Alters the row index of the true target within the context (random, first, or last position). \tabfms are designed to be permutation-invariant to row order; a model relying solely on ICL should produce identical predictions regardless of ordering. However, fine-tuning on static, fixed-order context-query pairs can introduce positional biases if optimization exploits ordering shortcuts. This manipulation tests whether any membership signal is feature-encoded in the model's weights or an artifact of positional over-indexing. 
    \item \textbf{Distribution Skewing:} Skews the marginal label distribution in the context. \emph{Skew true} over-represents the true target (amplification factor $\in \{2,3,5\}$); \emph{Skew incorrect} over-represents $k \in \{1,2,3\}$ incorrect targets (factor $\in \{2,3,5\}$); \emph{Mixed skewing} adjusts both simultaneously. \emph{Uniform and random scaling} amplify all candidate frequencies uniformly or randomly (factor $\in \{2,5,10\}$).
    \item \textbf{Feature Alteration:} \emph{Feature reduction} drops 2 valid feature columns to obscure pattern recognition; \emph{Distractor injection} appends 1--2 entirely random columns to introduce spurious signal. Both test whether the memorization signal survives changes to feature space.
    \item \textbf{Combined Stress Tests:} 2-way combinations pair positional shuffling with a single distributional or structural manipulation. 3-way stress tests simultaneously apply spatial, distributional, and structural degradation.
\end{enumerate}

\begin{table*}[htbp]
\centering
\caption{The complete suite of 48 adversarial context manipulations utilized to evaluate predictive robustness.}
\label{tab:perturbations}
\resizebox{\textwidth}{!}{%
\begin{tabular}{@{}llcp{23em}@{}}
\textbf{Perturbation Category} & \textbf{Varying Parameters} & \textbf{Count} & \textbf{Description} \\
\midrule
\textbf{Baseline} & N/A & 1 & Unperturbed, zero-information multiple-choice context. \\
\textbf{Positional Shuffling} & Position $\in \{\text{Random}, \text{First}, \text{Last}\}$ & 3 & Alters the spatial index of the true target row within the context sequence. \\
\textbf{Feature Reduction} & Columns removed $= 2$ & 1 & Drops valid feature columns to obscure pattern recognition. \\
\textbf{Distractor Injection} & Noisy columns $\in \{1, 2\}$ & 2 & Appends entirely random/noisy features to the context. \\
\textbf{Uniform Scaling} & Replication factor $\in \{2, 5, 10\}$ & 3 & Uniformly amplifies the frequency of all candidate labels in the context. \\
\textbf{Random Scaling} & Replication factor $\in \{2, 5, 10\}$ & 3 & Randomly alters the distribution frequencies of all candidate labels in the context. \\
\textbf{Skew True Target} & Amplification factor $\in \{2, 3, 5\}$ & 3 & Over-represents the true target label. \\
\textbf{Skew Incorrect Targets} & Targets skewed $k \in \{1, 2, 3\}$\newline Factor $\in \{2, 3, 5\}$ & 9 & Over-represents $k$ incorrect candidate labels. \\
\textbf{Mixed Skewing} & True factor $\in \{2, 3, 5\}$ Incorrect factor $\in \{2, 3, 5\}$ & 7 & Simultaneously skews both the true target and incorrect targets. \\
\textbf{2-Way Combinations} & Shuffle + \{Uniform, Skewed,\newline Random, Reduce\} & 9 & Combines positional shuffling with a single distributional or structural manipulation. \\
\textbf{3-Way Stress Tests} & Shuffle + Skew/Uniform/Random\newline + Feature Reduction & 7 & Stress tests simultaneously applying spatial, distributional, and structural degradation. \\
\midrule
\textbf{Total Configurations} & & \textbf{48} & \\
\end{tabular}%
}
\end{table*}

The count column in Tab.~\ref{tab:perturbations} specifies the number of distinct parameter configurations per category; the combinatorial categories (2-way, 3-way combinations) compose multiple single-axis manipulations to stress-test signal robustness simultaneously. Non-members rely entirely on the context for generalization and should exhibit clear loss spikes under strong degradation. Members, whose target mapping is encoded in the model's weights, should maintain a stable probe loss across all manipulations.

\para{Perturbation Effectiveness}
We evaluate perturbation effectiveness on the detectable memorization configurations, measuring the mean loss-based $\text{AUC}(\Delta S)$ per perturbation.
%
%Without any manipulation (baseline), the signal is already $\text{AUC}=0.543$, confirming that fine-tuned members and non-members diverge even under an unperturbed multiple-choice context.
%
\emph{Distribution skewing} is the most effective single perturbation: amplifying the true target label by $2\times$ yields AUC $=0.557$ ($\Delta{+}0.015$), and simultaneously amplifying all labels by $5\times$ yields the highest observed AUC of $0.559$ ($\Delta{+}0.017$).
This aligns with the design rationale: a memorized member resists context pressure because the correct label is encoded in its weights, whereas a non-member simply follows the over-represented label.
Also, as expected, \emph{Positional shuffling} is entirely ineffective (AUC $\approx 0.543$, $\Delta\approx 0$), confirming that memorization is stored as weight-level feature-label associations rather than positional heuristics, and that the model is indeed permutation-invariant to row order.
Feature reduction and distractor injection produce negligible gains (AUC $0.543$--$0.546$).
Combining positional shuffling with distribution skewing (\texttt{shuffle\_and\_skewed\_2}: AUC $=0.557$) does not improve over skewing alone, and heavily over-representing incorrect labels ($5\times$) slightly suppresses the signal (AUC $=0.539$), likely by reducing prediction confidence across all labels including the true one.
No individual perturbation amplifies the signal by more than $\Delta\text{AUC}\approx 0.017$; the practical value of the full perturbation suite lies in aggregating these stable but moderate individual signals into a more robust membership discriminator.
Future work could explore more aggressive perturbations (e.g., based on label cardinality or feature importance) to further amplify signal.

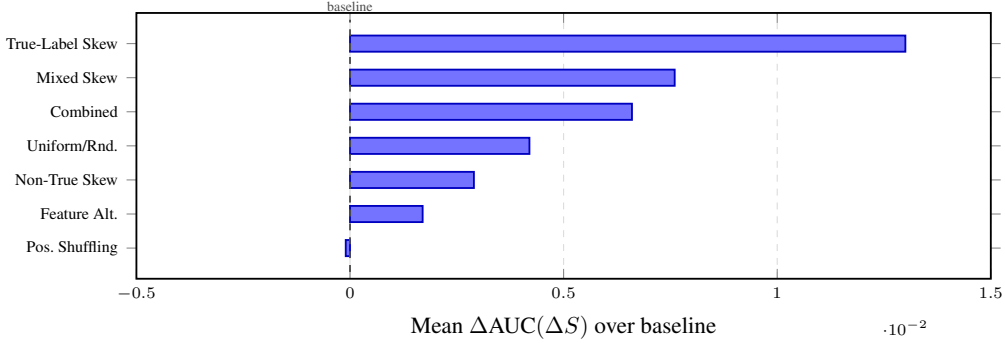
\begin{figure}[t]
\centering
\scalebox{0.87}{%
\begin{tikzpicture}
    \begin{axis}[
        xbar,
        bar width=0.7em,
        width=0.85\columnwidth,
        height=16em,
        xlabel={Mean $\Delta\text{AUC}(\Delta S)$ over baseline},
        xmin=-0.005, xmax=0.015,
        xtick={-0.005, 0, 0.005, 0.010, 0.015},
        xticklabel style={font=\scriptsize},
        ytick={1,2,3,4,5,6,7},
        yticklabels={Pos.\ Shuffling, Feature Alt., Non-True Skew, Uniform/Rnd., Combined, Mixed Skew, True-Label Skew},
        y tick label style={font=\scriptsize},
        xmajorgrids=true,
        grid style={dashed, gray!30},
        enlarge y limits=0.15,
        clip=false,
        thick,
    ]
    \addplot[fill=blue!55, draw=blue!75!black] coordinates {
        (-0.0001, 1)  % Positional Shuffling
        ( 0.0017, 2)  % Feature Alteration
        ( 0.0029, 3)  % Non-True Skewing
        ( 0.0042, 4)  % Uniform / Random
        ( 0.0066, 5)  % Combined
        ( 0.0076, 6)  % Mixed Skewing
        ( 0.0130, 7)  % True-Label Skewing
    };
    \draw[dashed, gray!60!black, line width=0.8pt]
        (axis cs:0, 0.3) -- (axis cs:0, 7.7)
        node[above, font=\tiny, gray!60!black] {baseline};
    \end{axis}
\end{tikzpicture}%
}
\vspace{-0.5em}
\caption{Mean $\Delta\text{AUC}(\Delta S)$ relative to the unperturbed baseline ($\text{AUC}_\text{baseline}{=}0.543$), per perturbation category averaged over detectable memorization configurations. True-label distribution skewing provides the strongest average lift ($\Delta{+}0.013$; best single perturbation: \texttt{mixed\_true\_5\_non\_true\_5}, $\Delta{+}0.017$). Positional shuffling is indistinguishable from baseline ($\Delta{\approx}0$).}
\label{fig:perturbation_effectiveness}
\vspace{-1em}
\end{figure}

\subsection{The Necessity of Difficulty Calibration}
\label{sec:appendix_calibration}

Raw, absolute probing metrics are unreliable for auditing memorization \cite{watson2022importance}. Pre-trained models already encode strong semantic priors and structural logic, achieving high prediction confidence on inherently easy datasets without task-specific fine-tuning. Relying on absolute metrics confounds this pre-existing structural bias with training-induced memorization.

\para{Impact on True Positive Rates (TPR)}
Across our \enx{530} structurally healthy (non-collapsed) configurations, raw loss evaluation flagged \enx{241} runs with uncalibrated $\text{TPR}_{@10\%} \ge 0.10$.
Difficulty calibration against $M_{ref}$ revealed that \enx{230} of these cases (\enx{95\%}) were purely data artifacts: the pre-trained model naturally extracted these records prior to fine-tuning due to inherent sample easiness. Applying the calibrated margin ($\Delta\text{TPR}_{@10\%} \ge 0.10$) reduces this to \enx{11} configurations with detectable memorization.

\para{Impact on AUC-ROC}
Global distributional metrics exhibit the same bias. If a dataset's features correlate strongly with a target label under the model's pre-training priors, $M_{ref}$ may yield a raw AUC of $0.85$ through zero-shot inference alone. An uncalibrated audit would flag a subsequent fine-tuned AUC of $0.87$ as a large signal. Computing delta scores $\Delta S(q) = S(q) - S_{ref}(q)$ and then $\text{AUC}(\Delta S)$ correctly identifies this as a structural artifact: since the fine-tuning contributes minimal additional discriminative power, $\text{AUC}(\Delta S) \approx 0.5$, indistinguishable from random guessing. Instance-level calibration via $\Delta S(q)$ is therefore necessary to isolate parametric memorization.

\section{Extended Evaluation Analysis}
\label{app:exteded_eval}

While Section~\ref{sec:eval} in the main text summarizes the five observed outcomes and their aggregate rates, this appendix provides a deeper analytical breakdown. We first detail the dominant artifact outcomes (representational collapse and loss inversion), which account for the majority of configurations, then analyze the secondary edge-case outcomes (context overfitting and zero-signal runs), track the temporal evolution of the memorization signal across fine-tuning epochs, and explore the disconnect between global and localized memorization.

\subsection{Artifact Outcomes: Collapse and Inversion}

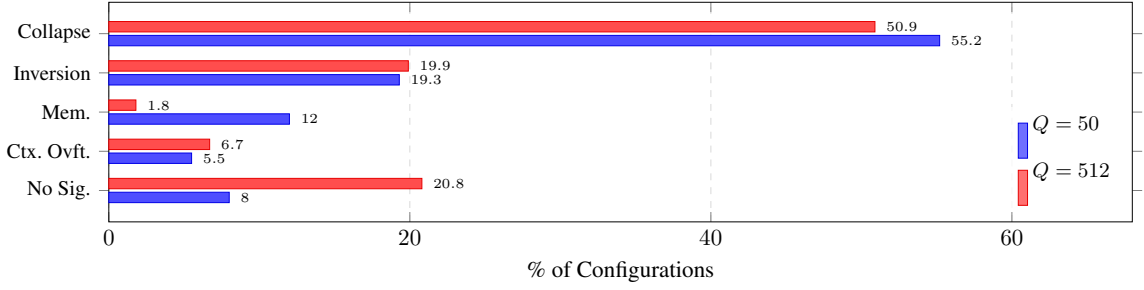
\begin{figure}[t]
\centering
\scalebox{0.87}{% adjust to fit available space
\begin{tikzpicture}
    \begin{axis}[
        xbar=0.15em,
        bar width=0.45em,
        width=\columnwidth,
        height=14em,
        enlarge y limits=0.2,
        legend style={
            at={(0.99,0.05)}, anchor=south east,
            draw=none, font=\small,
            legend cell align=left,
            fill=white, fill opacity=0.8, text opacity=1,
        },
        legend image code/.code={\draw[#1] (0,-0.08em) rectangle (0.4em,0.08em);},
        xlabel={\% of Configurations},
        xmin=0, xmax=68,
        xtick={0, 20, 40, 60},
        ytick={1,2,3,4,5},
        yticklabels={No Sig., Ctx.\ Ovft., Mem., Inversion, Collapse},
        y tick label style={font=\small},
        nodes near coords,
        nodes near coords align={horizontal},
        every node near coord/.append style={font=\tiny, xshift=0.15em},
        xmajorgrids=true,
        grid style={dashed, gray!30},
    ]
    \addplot[fill=blue!70, draw=blue!90!black] coordinates {
        (55.2,5) (19.3,4) (12.0,3) (5.5,2) (8.0,1)
    };
    \addplot[fill=red!70, draw=red!90!black] coordinates {
        (50.9,5) (19.9,4) (1.8,3) (6.7,2) (20.8,1)
    };
    \legend{$Q=50$, $Q=512$}
    \end{axis}
\end{tikzpicture}%
}% end \scalebox

\caption{Distribution of \iclmia outcomes. Abbreviations are: \emph{Mem}(orization), \emph{Ctx.\ Ovft.}: context overfitting; \emph{No Sig}(nal) }%Expanding the context to $Q=512$ nearly eliminates detectable memorization while shifting configurations toward no-signal outcomes.}
\label{fig:aggregate_outcomes}
\vspace{-1.5em}
\end{figure}

\para{Outcome I: Representational Collapse}
Affecting over half of our configurations (\enx{55.2\%} at $Q=50$; \enx{50.9\%} at $Q=512$), \emph{representational collapse} is the dominant fine-tuning artifact. At aggressive learning rates ($\lr \ge 10^{-2}$), \tabfms lose their in-context reasoning capability and output near-constant predictions. While this yields high uncalibrated AUCs ($> 0.7$), difficulty calibration confirms these are pre-existing dataset biases rather than induced memorization. We formally flag collapsed models using an empirically derived threshold: $\mathcal{L}_{FT} > 5 \times \mathcal{L}_{Base}$. Across our 1,128 runs, the loss ratio ($\mathcal{L}_{FT} / \mathcal{L}_{Base}$) follows a strict bimodal distribution. Stable configurations ($\lr \le 10^{-4}$) tightly cluster between $1.0$--$2.0\times$, whereas aggressive rates cause rapid degradation, pushing median ratios above $8.0\times$ (and frequently exceeding $10^4$). This $5\times$ boundary robustly separates structurally broken models from true parametric memorization.

\para{Outcome II: Loss Inversion}
In \enx{19.3\%} of configurations at $Q=50$, there is a further artifact: \emph{loss inversion}, where non-members yield lower mean probe loss than members after fine-tuning. This occurs when training loss is already inverted in the pre-trained model, or when the collapsed model's static predictions align better with non-members. % distribution.
Our framework addresses this structural bias by evaluating distributional metrics (i.e., entropy and confidence) which remain discriminative even when loss is inverted.

\subsection{Secondary Dynamics and Edge Cases}
\label{app:secondary_dynamics}
Our methodology isolates two additional outcomes beyond detectable memorization and representational collapse.

\para{Outcome IV: Context Overfitting}
In 5.5\% of runs at $Q=50$, the model learns statistical properties of the dataset schema rather than individual record membership. To classify this outcome, we define $g_k = \bar{\ell}^{M}_{\text{atk}}(M_\theta, \pi_k) - \bar{\ell}^{NM}_{\text{atk}}(M_\theta, \pi_k)$ as the mean member minus non-member probe loss gap under perturbation $\pi_k$, and let $\mu_{\text{gap}}$ and $\sigma_{\text{gap}}$ denote its mean and standard deviation across all 48 manipulations in $\Pi$. The mean probe loss gap ($\mu_{\text{gap}}$) is near zero, but $\sigma_{\text{gap}}$ is minimal across all 48 context manipulations: the zero is stable on every perturbation, not noisy. This indicates the model treats members and non-members identically under all context manipulations, having overfitted to the table format rather than individual rows.

\para{Outcome V: No Signal}
This outcome captures configurations (8.0\% at $Q=50$; 20.8\% at $Q=512$) where \iclmia produces no systematic pattern.
Noisy and inconsistent per-manipulation probe loss gaps ($g_k$) indicate that fine-tuning improved task performance without inducing a detectable row-level membership shift.

%\subsection{Temporal Evolution of Memorization}
%\input{figures/figure_memorization_evolution.tex}
%We track how the detectable memorization rate evolves over fine-tuning (Fig.~\ref{fig:epoch_evolution}). At $Q=50$, the signal starts at epoch 1 (\enx{2.7\%}), peaks around epoch 100 (\enx{8.2\%}), and reaches \enx{0\%} by epoch 500 as extended fine-tuning forces the model into representational collapse. At $Q=512$, memorization is substantially suppressed: it peaks at \enx{1.0\%} around epoch 50 and drops to \enx{0\%} at all subsequent checkpoints, confirming that larger query sizes not only reduce memorization rates but also accelerate its elimination.

%\subsection{Memorization vs.\ Exploitability}
\subsection{Overall memorization (AUC) vs confident instance signal (TPR at low FPR)}
\label{sec:appendix_vulnerable_datasets}

Our calibrated audit detected memorization ($\text{AUC}(\Delta S) > 0.5$) in 8 of the 10 evaluated tasks under specific hyperparameter regimes, despite representational collapse and chance predictions being the predominant outcomes.
Tab.~\ref{tab:exhaustive_vulnerability} reports the peak vulnerability configuration for each of these 8 tasks: learning rate, epoch, context size ($Q$), and the label randomization condition where the peak signal was detected. Normal indicates the true-label condition; Shuffled (M) indicates member-only label scrambling; Shuffled (All) indicates full-dataset scrambling. 
We report the calibrated $\text{AUC}(\Delta S)$ to capture global memorization trends across all records, complemented by $\text{TPR}$ at low $\text{FPR}$ ($\text{TPR}_{@10\%}$ and $\text{TPR}_{@1\%}$) to measure localized signal of individual samples. We report $AUC$ and $TPR$ for the metric (loss or entropy) yielding the maximum signal. 
% strength[cite: 278, 1032, 1050].

%We report the calibrated $\text{AUC}(\Delta S)$ as a global memorization metric, i.e., across all records, as well as the local measure targeting confidently inferring more individual samples - via TPR at low FPR, i.e., ($\text{TPR}_{@10\%}$ and $\text{TPR}_{@1\%}$). for the metric (loss or entropy) yielding the maximum signal. 

\begin{table*}[t!]
\centering
\caption{The peak detectability configurations for the 8 CARTE datasets, i.e., showing detectable memorization. The AUC and TPR metrics are computed directly on delta scores ($\Delta S$), where an AUC of 0.50 represents random guessing.}
\label{tab:exhaustive_vulnerability}
\resizebox{\textwidth}{!}{
\begin{tabular}{llccclrrr}
\textbf{Dataset} & \textbf{Label Condition} & \textbf{LR} & \textbf{Epoch} & \textbf{Q} & \textbf{Primary Metric} & \textbf{AUC} & \textbf{TPR$_{@10\%}$} & \textbf{TPR$_{@1\%}$} \\
\midrule
\texttt{babies\_r\_us} & Normal & $10^{-3}$ & 1 & 512 & Loss & 0.668 & 0.205 & 0.008 \\
\texttt{bikewale} & Shuffled (M) & $10^{-4}$ & 250 & 50 & Loss & 0.624 & 0.000 & 0.000 \\
\texttt{chocolate\_bar\_ratings} & Shuffled (M) & $10^{-2}$ & 100 & 50 & Loss & 0.611 & 0.180 & 0.020 \\
\texttt{buy\_buy\_baby} & Normal & $10^{-4}$ & 10 & 50 & Loss & 0.608 & 0.140 & 0.120 \\
\texttt{cardekho} & Normal & $10^{-4}$ & 1 & 50 & Loss & 0.596 & 0.140 & 0.000 \\
\texttt{beer\_ratings} & Shuffled (M) & $10^{-4}$ & 50 & 512 & Loss & 0.586 & 0.494 & 0.166 \\
\texttt{bikedekho} & Normal & $10^{-4}$ & 10 & 512 & Loss & 0.579 & 0.148 & 0.016 \\
\texttt{coffee\_ratings} & Normal & $10^{-3}$ & 10 & 50 & Loss & 0.560 & 0.200 & 0.000 \\
\end{tabular}
}
\end{table*}

\para{Global vs.\ Localized Memorization Signal}
Several configurations in Tab.~\ref{tab:exhaustive_vulnerability} yield a positive global memorization signal ($\text{AUC}(\Delta S) > 0.5$) while simultaneously exhibiting low TPR at strict FPR thresholds. Fine-tuning can induce a global distributional shift that separates members from non-members on average (AUC) without concentrating that signal at the individual-record level: $M_{ref}$ may achieve higher extreme-tail confidence on a subset of records purely through zero-shot structural bias. A detectable global signal therefore does not imply a strong localized one; row-level memorization is confined to a smaller subset of configurations under specific hyperparameter regimes.

\subsection{Data Characteristic Analysis}\label{sec:appendix_attribution}

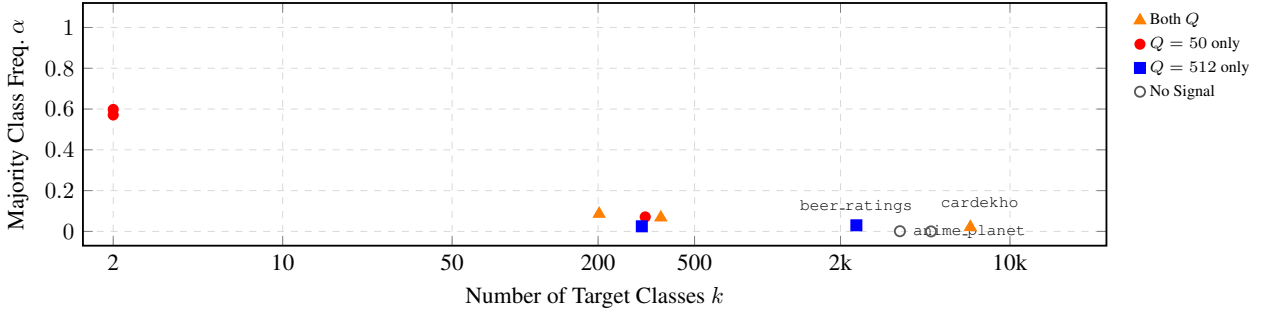
\begin{figure}[t]
\centering
\scalebox{0.87}{%
\begin{tikzpicture}
    \begin{axis}[
        width=\columnwidth,
        height=15em,
        xlabel={Number of Target Classes $k$},
        ylabel={Majority Class Freq.\ $\alpha$},
        xmode=log,
        xmin=1.5, xmax=25000,
        ymin=-0.07, ymax=1.12,
        xtick={2, 10, 50, 200, 500, 2000, 10000},
        xticklabels={2, 10, 50, 200, 500, 2k, 10k},
        ytick={0, 0.2, 0.4, 0.6, 0.8, 1.0},
        legend style={
            at={(1.02,1)}, anchor=north west,
            font=\scriptsize, draw=none,
            legend cell align=left,
        },
        grid=major,
        grid style={dashed, gray!30},
        thick,
        clip=false,
    ]

    % Both Q=50 and Q=512 (orange triangles)
    \addplot[only marks, mark=triangle*, color=orange, mark options={scale=1.3}]
        coordinates {
            (202, 0.086)  % babies_r_us
            (363, 0.068)  % bikewale
            (6865, 0.021) % cardekho
        };
    \addlegendentry{Both $Q$}

    % Q=50 only (red circles)
    \addplot[only marks, mark=*, color=red, mark options={scale=1.0}]
        coordinates {
            (2, 0.571)    % chocolate_bar_ratings
            (2, 0.598)    % coffee_ratings
            (313, 0.071)  % buy_buy_baby
        };
    \addlegendentry{$Q=50$ only}

    % Q=512 only (blue squares)
    \addplot[only marks, mark=square*, color=blue, mark options={scale=1.1}]
        coordinates {
            (303, 0.025)  % bikedekho
            (2325, 0.030) % beer_ratings
        };
    \addlegendentry{$Q=512$ only}

    % No signal (gray circles)
    \addplot[only marks, mark=o, color=gray!70!black, mark options={scale=1.1}]
        coordinates {
            (3516, 0.001) % anime_planet
            (4724, 0.000) % clear_corpus
        };
    \addlegendentry{No Signal}

    % Key annotations
    \node[font=\scriptsize, above=2pt] at (axis cs:2325, 0.030) {\texttt{beer\_ratings}};
    \node[font=\scriptsize, right=2pt] at (axis cs:3516, 0.001) {\texttt{anime\_planet}};
    \node[font=\scriptsize, above=5pt, xshift=4pt] at (axis cs:6865, 0.021) {\texttt{cardekho}};

    \end{axis}
\end{tikzpicture}%
}
\vspace{-0.5em}
\caption{Memorization signal category for each of the 10 evaluated CARTE datasets as a function of target cardinality $k$ and majority class frequency $\alpha$. The two points at $k{=}2$ represent \texttt{chocolate\_bar\_ratings} and \texttt{coffee\_ratings}; the two points near $k{=}300$ represent \texttt{bikedekho} (Q=512 only, $k=303$) and \texttt{buy\_buy\_baby} (Q=50 only, $k=313$).}
\label{fig:attribution_scatter}
\vspace{-1.5em}
\end{figure}

Fig.~\ref{fig:attribution_scatter} plots all 10 evaluated datasets in the target-cardinality--majority-class-frequency space, colored by memorization outcome category. Tab.~\ref{tab:attribution_datasets} reports the underlying statistics and per-$Q$ detectable memorization run counts.

\begin{table}[t]
\centering
\caption{Dataset statistics and detectable memorization run counts for all 10 evaluated CARTE tasks, sorted by target cardinality $k$. $n$: dataset size; $\alpha$: majority class frequency; Conf@50 / Conf@512: number of configurations yielding detectable memorization at $Q{=}50$ and $Q{=}512$, respectively.}
\label{tab:attribution_datasets}
\vspace{0.3em}
\scriptsize
\setlength{\tabcolsep}{4pt}
\begin{tabular}{lrrrcc}
\toprule
Dataset & $n$ & $k$ & $\alpha$ & Conf@50 & Conf@512 \\
\midrule
\texttt{chocolate\_bar} & 2.6k & 2 & 0.571 & 31 & 0 \\
\texttt{coffee\_ratings} & 2.1k & 2 & 0.598 & 1 & 0 \\
\midrule
\texttt{babies\_r\_us} & 5.1k & 202 & 0.086 & 1 & 3 \\
\texttt{bikedekho} & 4.8k & 303 & 0.025 & 0 & 2 \\
\texttt{buy\_buy\_baby} & 10.7k & 313 & 0.071 & 19 & 0 \\
\texttt{bikewale} & 9.0k & 363 & 0.068 & 14 & 1 \\
\midrule
\texttt{beer\_ratings} & 3.2k & 2325 & 0.030 & 0 & 3 \\
\texttt{anime\_planet} & 14.4k & 3516 & 0.001 & 0 & 0 \\
\texttt{clear\_corpus} & 4.7k & 4724 & 0.000 & 0 & 0 \\
\texttt{cardekho} & 37.8k & 6865 & 0.021 & 1 & 1 \\
\bottomrule
\end{tabular}
\end{table}

\para{Feature Composition}
CARTE datasets are mixed with both categorical and numerical features: \texttt{beer\_ratings} has 14 of 19 features as numeric, i.e., the highest ratio among all memorized datasets, producing near-unique feature vectors that ICL cannot generalize from, contributing to its parametric memorization at $Q=512$. In contrast, \texttt{clear\_corpus} (15/26 numeric) has similar richness but its continuous readability targets are feature-predictable, suppressing memorization. For predominantly categorical datasets, feature type does not discriminate memorized from non-memorized tasks; target cardinality and class concentration remain the primary factors.

\section{Exhaustive Hyperparameter Results}
\label{app:exhaustive_results}

Tabs.~\ref{tab:appendix_exhaustive_q50} and~\ref{tab:appendix_exhaustive_q512} report outcome metrics for $Q=50$ and $Q=512$ respectively, covering all 8 CARTE datasets and all evaluated learning rates. For each query size, we report the epoch yielding the highest signal of the reported outcome per learning rate \lr, selected independently for each $Q$.

\begin{table*}[th!]
\centering
\caption{Exhaustive evaluation of 8 CARTE datasets under \iclmia with $Q=50$ context queries. For each dataset, we track metrics across all four learning rates, selecting the peak-signal epoch independently for $Q=50$. $\Delta\mathcal{L}_v$: vanilla loss delta $\mathcal{L}_v(M_\theta) - \mathcal{L}_v(M_{ref})$ for members (IN) and non-members (OUT). \textbf{N}: normal label condition; \textbf{S}: member-shuffled condition. $\text{AUC}(\Delta S)$ and TPR are computed on instance-level delta scores $\Delta S(q) = S(q) - S_{\text{ref}}(q)$; bold AUC = detectable memorization ($\text{AUC}(\Delta S) > 0.5$). \textbf{Pert.}: mean $\mu_{\text{gap}}$ and std.\ $\sigma_{\text{gap}}$ of the per-manipulation member--non-member probe loss gap $g_k$ across all $\pi_k \in \Pi$. \textbf{Outcome}: \textbf{Mem.}~=~Detectable Memorization, Coll.~=~Collapse, Inv.~=~Inversion, Ctx.Ov.~=~Context Overfitting, No Sig.~=~No Signal.}
\label{tab:appendix_exhaustive_q50}
\scriptsize
\setlength{\tabcolsep}{0.2em}
\begin{tabular}{llcrrrrrrrrl}
 & & & \multicolumn{2}{c}{\textbf{$\Delta\mathcal{L}_v$}} & \multicolumn{2}{c}{{\textbf{$\text{AUC}(\Delta S)$}}} & \multicolumn{2}{c}{\textbf{Pert.}} & \multicolumn{2}{c}{{\textbf{TPR}}} & \\
\cmidrule(lr){4-5} \cmidrule(lr){6-7} \cmidrule(lr){8-9} \cmidrule(lr){10-11}
\textbf{Dataset} & \textbf{LR} & \textbf{Ep.}
  & \textbf{IN} & \textbf{OUT}
  & \textbf{N} & \textbf{S}
  & $\mu_{\text{gap}}$ & $\sigma_{\text{gap}}$
  & \textbf{@10\%} & \textbf{@1\%}
  & \textbf{Outcome} \\
\midrule
\multirow{4}{*}{\texttt{babies\_r\_us}}
 & $10^{-1}$ & 10  & $1.9{\cdot}10^{7}$ & $1.4{\cdot}10^{7}$ & 1.000 & 1.000 & $-3.7{\cdot}10^{5}$ & $9.2{\cdot}10^{4}$ & 1.00 & 1.00 & Coll. \\
 & $10^{-2}$ & 50  & $9.1{\cdot}10^{4}$ & $6.8{\cdot}10^{4}$ & 0.997 & 0.692 & $-1.9{\cdot}10^{3}$ & 573.4 & 0.98 & 0.98 & Coll. \\
 & $10^{-3}$ & 1  & 0.502 & 0.578 & 0.462 & 0.464 & -0.148 & 0.737 & 0.04 & 0 & No Sig. \\
 & $10^{-4}$ & 50  & -0.184 & -0.034 & $\mathbf{0.665}$ & 0.456 & -0.161 & 0.724 & 0.02 & 0 & \textbf{Mem.} \\
\midrule
\multirow{4}{*}{\texttt{bikewale}}
 & $10^{-1}$ & 1  & $2.8{\cdot}10^{4}$ & $2.2{\cdot}10^{4}$ & 0.374 & 0.481 & 20.47 & 78.79 & 0.02 & 0 & Coll. \\
 & $10^{-2}$ & 50  & $7.4{\cdot}10^{4}$ & $5.8{\cdot}10^{4}$ & 0.350 & 0.597 & 55.54 & 168.7 & 0.02 & 0 & Coll. \\
 & $10^{-3}$ & 1  & 0.734 & 0.959 & 0.518 & 0.486 & -0.039 & 0.333 & 0.08 & 0.04 & Coll. \\
 & $10^{-4}$ & 50  & -0.052 & 0.009 & $\mathbf{0.578}$ & $\mathbf{0.563}$ & -0.017 & 0.156 & 0 & 0 & \textbf{Mem.} \\
\midrule
\multirow{4}{*}{\texttt{chocolate\_bar\_ratings}}
 & $10^{-1}$ & 50  & -0.024 & 0.165 & $\mathbf{0.585}$ & $\mathbf{0.594}$ & -0.007 & 0.063 & 0.16 & 0 & \textbf{Mem.} \\
 & $10^{-2}$ & 50  & -0.010 & 0.145 & $\mathbf{0.604}$ & $\mathbf{0.605}$ & 0.004 & 0.041 & 0.16 & 0 & \textbf{Mem.} \\
 & $10^{-3}$ & 10  & 0.021 & 0.162 & $\mathbf{0.609}$ & $\mathbf{0.606}$ & 0.009 & 0.085 & 0.18 & 0.02 & \textbf{Mem.} \\
 & $10^{-4}$ & 10  & 0.099 & 0.039 & 0.466 & $\mathbf{0.589}$ & 0.054 & 0.310 & 0.10 & 0.02 & \textbf{Mem.} \\
\midrule
\multirow{4}{*}{\texttt{buy\_buy\_baby}}
 & $10^{-1}$ & 1  & $9.2{\cdot}10^{3}$ & $5.9{\cdot}10^{3}$ & 0.998 & 1.000 & -761.1 & 313.9 & 1.00 & 0.96 & Coll. \\
 & $10^{-2}$ & 10  & $3.8{\cdot}10^{4}$ & $2.4{\cdot}10^{4}$ & 1.000 & 1.000 & $-2.8{\cdot}10^{3}$ & 722.2 & 1.00 & 1.00 & Coll. \\
 & $10^{-3}$ & 1  & 0.452 & 0.269 & $\mathbf{0.553}$ & $\mathbf{0.563}$ & -0.307 & 1.871 & 0.10 & 0 & \textbf{Mem.} \\
 & $10^{-4}$ & 10  & -0.060 & -0.011 & $\mathbf{0.608}$ & $\mathbf{0.577}$ & -0.213 & 1.791 & 0.14 & 0.12 & \textbf{Mem.} \\
\midrule
\multirow{4}{*}{\texttt{cardekho}}
 & $10^{-1}$ & 1  & 356.1 & 537.5 & 0.351 & 0.041 & 0.902 & 2.084 & 0.04 & 0 & Coll. \\
 & $10^{-2}$ & 1  & 7.705 & 12.75 & 0.546 & 0.540 & -0.026 & 0.324 & 0.24 & 0.06 & Coll. \\
 & $10^{-3}$ & 1  & 0.808 & 0.864 & 0.558 & 0.556 & 0.009 & 0.070 & 0.24 & 0.06 & Coll. \\
 & $10^{-4}$ & 1  & 0.010 & 0.028 & $\mathbf{0.596}$ & 0.497 & 0.011 & 0.067 & 0.14 & 0 & \textbf{Mem.} \\
\midrule
\multirow{4}{*}{\texttt{beer\_ratings}}
 & $10^{-1}$ & 1  & $3.3{\cdot}10^{4}$ & $2.8{\cdot}10^{4}$ & 0.019 & 0.294 & 286.4 & 119.2 & 0 & 0 & Coll. \\
 & $10^{-2}$ & 1  & 5.249 & 4.765 & 0.480 & 0.511 & -0.003 & 1.558 & 0.08 & 0.04 & Coll. \\
 & $10^{-3}$ & 1  & 1.462 & 1.265 & 0.481 & 0.495 & -0.007 & 0.460 & 0.10 & 0 & No Sig. \\
 & $10^{-4}$ & 10  & -0.198 & 0.121 & 0.507 & 0.509 & -0.039 & 0.530 & 0.08 & 0.02 & No Sig. \\
\midrule
\multirow{4}{*}{\texttt{bikedekho}}
 & $10^{-1}$ & 1  & $3.7{\cdot}10^{3}$ & $3.0{\cdot}10^{3}$ & --- & --- & 57.83 & 16.07 & 0 & 0 & Coll. \\
 & $10^{-2}$ & 1  & 2.986 & 1.705 & 0.428 & 0.436 & 0.057 & 0.361 & 0.08 & 0 & Coll. \\
 & $10^{-3}$ & 1  & 4.262 & 3.423 & 0.537 & 0.532 & -0.035 & 0.236 & 0.10 & 0.06 & Ctx.Ov. \\
 & $10^{-4}$ & 10  & -0.005 & 0.018 & 0.542 & 0.530 & -0.005 & 0.117 & 0.14 & 0.06 & Ctx.Ov. \\
\midrule
\multirow{4}{*}{\texttt{coffee\_ratings}}
 & $10^{-1}$ & 10  & 9.375 & 9.335 & 0.498 & 0.519 & 0.113 & 2.717 & 0.08 & 0 & Coll. \\
 & $10^{-2}$ & 50  & 0.307 & 0.267 & 0.528 & 0.536 & 0.004 & 0.106 & 0.08 & 0.02 & Inv. \\
 & $10^{-3}$ & 10  & 0.177 & 0.137 & $\mathbf{0.560}$ & 0.508 & -0.006 & 0.153 & 0.20 & 0 & \textbf{Mem.} \\
 & $10^{-4}$ & 50  & 0.182 & 0.292 & 0.520 & 0.529 & 0.011 & 0.173 & 0.08 & 0.02 & Inv. \\
\bottomrule
\end{tabular}
\end{table*}

\begin{table*}[th!]
\centering
\caption{Exhaustive evaluation of 8 CARTE datasets under \iclmia with $Q=512$ context queries. For each dataset, we track metrics across all four learning rates, selecting the peak-signal epoch independently for $Q=512$. $\Delta\mathcal{L}_v$: vanilla loss delta $\mathcal{L}_v(M_\theta) - \mathcal{L}_v(M_{ref})$ for members (IN) and non-members (OUT). \textbf{N}: normal label condition; \textbf{S}: member-shuffled condition. $\text{AUC}(\Delta S)$ and TPR are computed on instance-level delta scores $\Delta S(q) = S(q) - S_{\text{ref}}(q)$; bold AUC = detectable memorization ($\text{AUC}(\Delta S) > 0.5$). \textbf{Pert.}: mean $\mu_{\text{gap}}$ and std.\ $\sigma_{\text{gap}}$ of the per-manipulation member--non-member probe loss gap $g_k$ across all $\pi_k \in \Pi$. \textbf{Outcome}: \textbf{Mem.}~=~Detectable Memorization, Coll.~=~Collapse, Inv.~=~Inversion, Ctx.Ov.~=~Context Overfitting, No Sig.~=~No Signal.}
\label{tab:appendix_exhaustive_q512}
\scriptsize
\setlength{\tabcolsep}{0.2em}
\begin{tabular}{llcrrrrrrrrl}
 & & & \multicolumn{2}{c}{\textbf{$\Delta\mathcal{L}_v$}} & \multicolumn{2}{c}{{\textbf{$\text{AUC}(\Delta S)$}}} & \multicolumn{2}{c}{\textbf{Pert.}} & \multicolumn{2}{c}{{\textbf{TPR}}} & \\
\cmidrule(lr){4-5} \cmidrule(lr){6-7} \cmidrule(lr){8-9} \cmidrule(lr){10-11}
\textbf{Dataset} & \textbf{LR} & \textbf{Ep.}
  & \textbf{IN} & \textbf{OUT}
  & \textbf{N} & \textbf{S}
  & $\mu_{\text{gap}}$ & $\sigma_{\text{gap}}$
  & \textbf{@10\%} & \textbf{@1\%}
  & \textbf{Outcome} \\
\midrule
\multirow{4}{*}{\texttt{babies\_r\_us}}
 & $10^{-1}$ & 1  & 206.1 & 171.3 & 0.276 & --- & 18.09 & 31.07 & 0.02 & 0 & Coll. \\
 & $10^{-2}$ & 1  & 0.711 & 0.742 & 0.487 & 0.463 & -0.075 & 1.234 & 0.12 & 0.01 & Coll. \\
 & $10^{-3}$ & 1  & 0.569 & 0.535 & $\mathbf{0.668}$ & --- & -0.102 & 1.285 & 0.21 & 0.01 & \textbf{Mem.} \\
 & $10^{-4}$ & 10  & -0.033 & 0.004 & $\mathbf{0.608}$ & 0.482 & -0.085 & 1.289 & 0.02 & 0 & \textbf{Mem.} \\
\midrule
\multirow{4}{*}{\texttt{bikewale}}
 & $10^{-1}$ & 1  & $3.0{\cdot}10^{4}$ & $2.7{\cdot}10^{4}$ & 0.193 & 0.121 & 69.14 & 83.08 & 0.00 & 0 & Coll. \\
 & $10^{-2}$ & 1  & 49.04 & 42.88 & 0.485 & 0.466 & 0.084 & 2.658 & 0.09 & 0.01 & Coll. \\
 & $10^{-3}$ & 1  & 0.408 & 0.412 & $\mathbf{0.582}$ & --- & -0.009 & 0.264 & 0.27 & 0.12 & \textbf{Mem.} \\
 & $10^{-4}$ & 10  & -0.021 & 0.008 & 0.536 & 0.521 & -0.009 & 0.264 & 0.11 & 0.01 & No Sig. \\
\midrule
\multirow{4}{*}{\texttt{chocolate\_bar\_ratings}}
 & $10^{-1}$ & 50  & 2.677 & 1.956 & 0.496 & 0.511 & 0.008 & 1.122 & 0.09 & 0.01 & No Sig. \\
 & $10^{-2}$ & 50  & 0.067 & 0.025 & 0.514 & 0.516 & 0.000 & 0.056 & 0.10 & 0.02 & Inv. \\
 & $10^{-3}$ & 10  & 0.092 & 0.200 & 0.512 & --- & -0.001 & 0.175 & 0.10 & 0.02 & Ctx.Ov. \\
 & $10^{-4}$ & 10  & 0.009 & -0.012 & 0.516 & 0.510 & -0.000 & 0.055 & 0.09 & 0.01 & Ctx.Ov. \\
\midrule
\multirow{4}{*}{\texttt{buy\_buy\_baby}}
 & $10^{-1}$ & 1  & $1.5{\cdot}10^{4}$ & $1.7{\cdot}10^{4}$ & --- & --- & $2.5{\cdot}10^{3}$ & 686.5 & 0 & 0 & Coll. \\
 & $10^{-2}$ & 1  & 19.78 & 27.81 & 0.431 & 0.424 & 6.079 & 35.42 & 0.05 & 0 & Coll. \\
 & $10^{-3}$ & 50  & 0.764 & 0.741 & 0.513 & --- & -0.069 & 2.860 & 0.14 & 0.03 & Coll. \\
 & $10^{-4}$ & 50  & -0.024 & -0.002 & 0.461 & 0.521 & -0.002 & 2.326 & 0.11 & 0.02 & No Sig. \\
\midrule
\multirow{4}{*}{\texttt{cardekho}}
 & $10^{-1}$ & 10  & $5.7{\cdot}10^{7}$ & $6.0{\cdot}10^{7}$ & 0.464 & 0.536 & $1.2{\cdot}10^{3}$ & $1.5{\cdot}10^{4}$ & 0.08 & 0.02 & Coll. \\
 & $10^{-2}$ & 10  & $2.0{\cdot}10^{3}$ & $2.1{\cdot}10^{3}$ & 0.552 & 0.492 & -0.455 & 4.224 & 0.11 & 0.02 & Coll. \\
 & $10^{-3}$ & 50  & 250.3 & 261.8 & 0.489 & --- & 0.021 & 1.473 & 0.10 & 0.02 & Coll. \\
 & $10^{-4}$ & 50  & 0.043 & 0.056 & $\mathbf{0.560}$ & 0.489 & 0.001 & 0.066 & 0.03 & 0.00 & \textbf{Mem.} \\
\midrule
\multirow{4}{*}{\texttt{beer\_ratings}}
 & $10^{-1}$ & 10  & $2.3{\cdot}10^{7}$ & $2.2{\cdot}10^{7}$ & 0.999 & 0.909 & $-1.9{\cdot}10^{5}$ & $5.1{\cdot}10^{4}$ & 1.00 & 1.00 & Coll. \\
 & $10^{-2}$ & 50  & $2.6{\cdot}10^{3}$ & $2.5{\cdot}10^{3}$ & 0.792 & 0.996 & -19.68 & 26.75 & 0.24 & 0.04 & Coll. \\
 & $10^{-3}$ & 1  & 0.492 & 0.470 & 0.502 & --- & -0.034 & 0.619 & 0.08 & 0.01 & No Sig. \\
 & $10^{-4}$ & 50  & -0.009 & 0.026 & 0.497 & $\mathbf{0.586}$ & -0.033 & 0.578 & 0.49 & 0.17 & \textbf{Mem.} \\
\midrule
\multirow{4}{*}{\texttt{bikedekho}}
 & $10^{-1}$ & 1  & $1.0{\cdot}10^{4}$ & $9.9{\cdot}10^{3}$ & --- & --- & 218.2 & 32.48 & 0 & 0 & Coll. \\
 & $10^{-2}$ & 1  & 5.159 & 4.631 & 0.442 & 0.473 & 0.101 & 0.643 & 0.07 & 0.01 & Coll. \\
 & $10^{-3}$ & 1  & 1.029 & 1.036 & 0.473 & --- & 0.013 & 0.228 & 0.13 & 0.02 & Coll. \\
 & $10^{-4}$ & 10  & -0.025 & 0.007 & $\mathbf{0.579}$ & 0.497 & 0.006 & 0.190 & 0.15 & 0.02 & \textbf{Mem.} \\
\midrule
\multirow{4}{*}{\texttt{coffee\_ratings}}
 & $10^{-1}$ & 50  & 0.284 & 0.246 & 0.519 & 0.517 & -0.004 & 0.156 & 0.09 & 0.00 & Ctx.Ov. \\
 & $10^{-2}$ & 50  & 0.263 & 0.223 & 0.519 & 0.522 & -0.003 & 0.142 & 0.09 & 0.00 & Ctx.Ov. \\
 & $10^{-3}$ & 10  & 0.178 & 0.134 & 0.521 & --- & -0.002 & 0.074 & 0.10 & 0.00 & Ctx.Ov. \\
 & $10^{-4}$ & 50  & 0.275 & 0.128 & 0.522 & 0.525 & -0.001 & 0.079 & 0.10 & 0.01 & Ctx.Ov. \\
\bottomrule
\end{tabular}
\end{table*}

\section{Memorization under Realistic Training Regimes}\label{app:ablation_realistic}

% Table: Ablation — Combined Training vs. Single-Task, Q=512 (signal-only datasets)
\begin{table*}[t]
\centering
\caption{Peak detectable memorization at $Q{=}512$, for datasets with signal in at least one configuration. \emph{Single-Task Fixed Pairs}: one model fine-tuned per dataset (peak across all $\lr$; $^\star$$\lr=10^{-3}$, others $\lr=10^{-4}$). \emph{Multi-Task Fixed Pairs}: one model trained jointly on all 10 CARTE tasks with fixed context-query pairs ($\lr=10^{-4}$). \emph{Multi-Task Random Pairs}: same joint setup with a fresh random context-query sample at each step ($\lr=10^{-4}$). \textbf{Bold}: highest AUC per row.}
\label{tab:ablation_q512}
\scriptsize
\setlength{\tabcolsep}{0.38em}
\begin{tabular}{l rrrr rrrr rrrr}
\toprule
 & \multicolumn{4}{c}{Single-Task Fixed Pairs} & \multicolumn{4}{c}{Multi-Task Fixed Pairs} & \multicolumn{4}{c}{Multi-Task Random Pairs} \\
\cmidrule(lr){2-5}\cmidrule(lr){6-9}\cmidrule(lr){10-13}
\textbf{Dataset} & Ep & AUC & $\text{TPR}_{@10\%}$ & $\text{TPR}_{@1\%}$ & Ep & AUC & $\text{TPR}_{@10\%}$ & $\text{TPR}_{@1\%}$ & Ep & AUC & $\text{TPR}_{@10\%}$ & $\text{TPR}_{@1\%}$ \\
\midrule
\texttt{babies\_r\_us} & 1 & $^\star$\textbf{0.668} & 0.205 & 0.008 & 100 & 0.618 & 0.008 & 0.000 & \multicolumn{4}{c}{---} \\
\texttt{beer\_ratings}  & 50 & \textbf{0.586} & 0.494 & 0.166 & \multicolumn{4}{c}{---} & \multicolumn{4}{c}{---} \\
\texttt{bikedekho}     & 10 & \textbf{0.579} & 0.148 & 0.016 & 1   & 0.558 & 0.190 & 0.027 & \multicolumn{4}{c}{---} \\
\texttt{bikewale}      & 1  & $^\star$\textbf{0.582} & 0.266 & 0.123 & \multicolumn{4}{c}{---} & 10 & 0.575 & 0.100 & 0.018 \\
\texttt{cardekho}      & 50 & \textbf{0.560} & 0.029 & 0.002 & \multicolumn{4}{c}{---} & \multicolumn{4}{c}{---} \\
\bottomrule
\end{tabular}
\end{table*}

\para{Memorization Signal under Realistic Regimes ($Q{=}512$)}
Tab.~\ref{tab:ablation_q512} shows how memorization degrades as the training regime progressively approaches practical conditions. 
Under single-task fine-tuning on fixed context-query pairs (worst case), 5 of 10 tasks are detectable.
By fine-tuning a single model jointly on all 10 tasks with fixed context-query pairs, i.e., 10 gradient steps per epoch, the signal appears only for 2 tasks (\texttt{babies\_r\_us}: AUC 0.668$\to$0.618; \texttt{bikedekho}: AUC 0.579$\to$0.558).
We intentionally fixed the context-query pairs across epochs, however, in practice the pairs are randomly sampled.
With random context-query sampling and fine-tuning on all 10 tasks, the signal collapses to a single task: \texttt{bikewale} (AUC 0.575).
Critically, this residual signal appears only at Epoch~10, already beyond the 2--5 epoch budget of real \tabfms pre-training~\cite{spinaci2025contexttab}; within that practical budget, no task would be detectable.
The pattern confirms that memorization under \iclmia requires two structural conditions absent from practical pre-training: (i)~per-task gradient specialization and (ii)~repeated exposure to fixed context-query pairs.

\end{document}